\documentclass{article}

    \PassOptionsToPackage{numbers, compress}{natbib}
\usepackage[final]{neurips_2021}

\usepackage[utf8]{inputenc} % allow utf-8 input
\usepackage[T1]{fontenc}    % use 8-bit T1 fonts
\usepackage[draft]{hyperref}       % hyperlinks             % TODO - remove [draft], which was added to avoid a compilation issue
\usepackage{url}            % simple URL typesetting
\usepackage{booktabs}       % professional-quality tables
\usepackage{amsfonts}       % blackboard math symbols
\usepackage{nicefrac}       % compact symbols for 1/2, etc.
\usepackage{microtype}      % microtypography
\usepackage{xcolor}         % colors
\usepackage{caption}
\usepackage[flushleft]{threeparttable}
\usepackage{longtable}

\usepackage{soul}           % for highlighting
\usepackage{graphicx}
\usepackage{float}

\title{Evaluation of Human-AI Teams for Learned and Rule-Based Agents in Hanabi}

\newcommand*\samethanks[1][\value{footnote}]{\footnotemark[#1]}

\author{%
  Ho Chit Siu \thanks{MIT Lincoln Laboratory, \texttt{\{hochit.siu,jdpena,yutai.zhou,chestnut,ross.allen\}@ll.mit.edu}}\And
  Jaime D. Pe\~na \samethanks \And
  Yutai Zhou \samethanks[1] \And
  Edenna Chen \thanks{MIT Department of Electrical Engineering and Computer Science, \texttt{edenna@mit.edu}} \And
  Victor J. Lopez \thanks{U.S. Air Force Artificial Intelligence Accelerator, \texttt{\{victor.lopez.10,kyle.palko.1\}@us.af.mil}} \And
  Kyle Palko  \samethanks \And
  Kimberlee C. Chang \samethanks[1] \And
  Ross E. Allen \samethanks[1]
}

\begin{document}

\maketitle

\begin{abstract}
Deep reinforcement learning has generated superhuman AI in competitive games such as Go and StarCraft. Can similar learning techniques create a superior AI teammate for human-machine collaborative games? Will humans prefer AI teammates that improve objective team performance or those that improve subjective metrics of trust? In this study, we perform a single-blind evaluation of teams of humans and AI agents in the cooperative card game \emph{Hanabi}, with both rule-based and learning-based agents. In addition to the game score, used as an objective metric of the human-AI team performance, we also quantify subjective measures of the human's perceived performance, teamwork, interpretability, trust, and overall preference of AI teammate. We find that humans have a clear preference toward a rule-based AI teammate (SmartBot) over a state-of-the-art learning-based AI teammate (Other-Play) across nearly all subjective metrics, and generally view the learning-based agent negatively, despite no statistical difference in the game score. This result has implications for future AI design and reinforcement learning benchmarking, highlighting the need to incorporate subjective metrics of human-AI teaming rather than a singular focus on objective task performance.
\footnote{DISTRIBUTION STATEMENT A. Approved for public release. Distribution is unlimited.}\end{abstract}

\section{Introduction}

Advances in artificial intelligence (AI) have resulted in agents that perform at superhuman levels within domains previously thought to be solely the purview of human intelligence. Such performance has been demonstrated through application of reinforcement learning (RL) in environments such as board games \citep{campbell2002deep,silver2018general}, arcade games \citep{mnih2013playing}, real-time strategy games \cite{vinyals2019grandmaster}, multiplayer online battle arenas \citep{berner2019dota}, and simulated aerial dogfights \citep{alphadogfight}.

In nearly all cases, these demonstrations of AI superiority are in purely adversarial, one- or two-player games. %However, in order for such systems to have real-world applicability, a major element of \emph{teaming intelligence} must also be shown, particularly with human teammates \citep{johnson2019no}. %
However, in order to achieve real-world applicability and adoption, AI must be able to demonstrate \emph{teaming intelligence}, particularly with human teammates \citep{johnson2019no}.
Due to the focus on adversarial games, teaming intelligence has been understudied in RL research. % means that teaming intelligence has been understudied for systems using the kind of reinforcement learning (RL). %that underpins much of the aforementioned success. 
AI teammates must also exhibit behavior that engenders an appropriate level of certain human reactions, such as trust, mental workload, and risk perception \citep{lee2004trust,parasuraman1997humans}. Failure to do so risks the same kind of misuse, disuse, and abuse that Parasuraman and Riley illustrated with traditional automation systems \citep{parasuraman1997humans}. %among other barriers to practical implementation. 
These issues are distinct from much of current multi-agent AI work, as AI that is able to team effectively with other AI agents has failed to work effectively with humans \citep{carroll2019utility}. %Issues with human use of automation have been particularly well-studied in aerospace human factors \citep{dzindolet2001predicting,tsang2002principles,mindell2011digital,dehais2015automation}, but modern notions of human-\emph{AI} teaming suffer from many of the same deficiencies and a lack of research \citep{johnson2019no}.

The objective of this paper is to evaluate human-AI teaming in the cooperative, imperfect-information game of Hanabi. We consider not only the \emph{objective} performance of a human-AI team, but also the \emph{subjective} human reactions and preferences when working with different kinds of AI teammates. %
Based on the success of applying deep reinforcement learning to create superhuman AI in adversarial games, we hypothesize that similar RL techniques can render collaborative AI that outperform and are preferred over rule-based agents in human-AI Hanabi teams. 
Our results show that this hypothesis is \emph{not} supported given the current state of the art of collaborative RL agents. %
%Our results indicate that 
Human participants show a clear preference toward rule-based AI even though the learning-based AI  perform no worse and are specifically optimized for teaming with previously unknown partners (e.g. humans) \cite{hu2020other}. 
To the best of our knowledge, this is the first comparative study of objective performance of rule-based and learning-based Hanabi AI in human-teaming experiments, as well as the first quantified study of subjective human preferences toward such AI. %

% In Section \ref{sec:background} we present some background on our Hanabi experiment environment, AI approaches for Hanabi, and some human-AI teaming metrics. In Section \ref{sec:methods} we describe the experimental procedure, along with the metrics used to evaluate the teams. Sections \ref{sec:results} details experimental results. Section \ref{sec:discussion} discusses some implications of these results and broader applications to training agents for human-AI teams.

\section{Background}
\label{sec:background}

\subsection{Hanabi}
\label{subsection:hanabi}
\emph{Hanabi} is a cooperative card game in which two to five players attempt to stack twenty-five cards into five different fireworks (piles), one for each suit (color) and by ascending rank (number). We consider only the standard, two-player version, where the deck is composed of 50 cards, five suits, each suit having three 1s, two 2s, two 3s, two 4s, and one 5, and each player is dealt a hand of five cards. Hanabi’s difficulty lies in the fact that players can only see their teammate’s hand and never their own and communication about cards is strictly limited.

Games start with eight \emph{hint tokens} and three \emph{bomb tokens}. Each turn, a player may \emph{discard} a card from their hand, \emph{play} a card from their hand, or give a \emph{hint} to a teammate. Whenever a card is played or discarded, it is revealed and a new card is drawn from the deck. A correctly-played card is placed in the appropriate firework; an incorrectly-played card is discarded and the team loses one bomb. Hinting costs one hint token and allows a player to reveal either all cards of a certain suit or rank in their teammate’s hand. Hint tokens are earned back when a card is discarded, or when a 5 card is successfully played. The final score is the sum of the top card in each firework, for a maximum of 25 points. The game ends when all fireworks have been completed, the deck is empty (and each player has one additional turn), or the team loses 3 bombs.

Hanabi is a \textit{purely cooperative} game with imperfect information, and limited, codified communication. These properties make it an interesting challenge for teaming since players must consider the reasons for their teammates' actions and any implied information, while avoiding misinterpretations. \citet{bard2020hanabi} presents a more complete treatment of Hanabi and its properties as an AI problem.

\subsection{Training and Evaluating AI Teams}
Hanabi's uniquely collaborative nature has made it the subject of several AI challenges in recent years \citep{bard2020hanabi,walton20192018}. %
In these challenges AI agents are paired with other teammate agents, and their performance is measured based on the paradigms of \textit{self-play} and \textit{cross-play} \citep{hu2020other}. %also referred to as mirror-play and mixed-play, respectively \citep{canaan2020generating}. 

Self-play (SP) is when an AI plays a game with a copy of itself as the opponent (adversarial games) or teammate (cooperative games like Hanabi). Self-play can be used as a form of RL training \citep{bansal2017emergent,hu2019simplified}, and/or as a form of evaluation  \citep{bard2020hanabi}. %The two meanings are closely related but are not synonymous; we will attempt to disambiguate the two throughout this paper. %In the context of AI evaluation, self-play is also referred to as \textit{mirror-play}.

Cross-play (XP) is an evaluation-only paradigm where an agent is teamed with other agents (AI or humans) that were not encountered during training, measuring how well an agent can cooperate with previously-unseen teammates. %
SP-trained agents can achieve high SP-scores by developing ``secretive conventions'' that are not understood by agents not present during training, thus completely failing in the XP setting \citep{hu2020other}.
Cross-play with humans (\emph{human-play}) is of particular importance as it measures human-machine teaming (HMT) and is the foundation for the experiments in our paper. %Cross-play is also referred to as \textit{mixed-play} \citep{canaan2020generating}.

Cross-play can be evaluated under the zero-shot coordination (ZSC) \citep{hu2020other} or ad-hoc teaming \citep{Stone2010AdHA} settings. ZSC assumes that teams are formed between agents that have no prior knowledge of each other; therefore it is impossible to train an agent to bias towards teammate idiosyncrasies, though the agents may agree to a common training strategy beforehand. On the other hand, ad-hoc teaming attempts to achieve coordination by having agents update their policy while interacting with other agents, and the the pool of partner agents can be known in advance (though it can be of arbitrarily large size, making such knowledge practically impossible to exploit).
% XP was developed to measure progress towards zero-shot coordination (ZSC): constructing AI agents that can coordinate with previously-unseen partners. While closely related to ad-hoc teaming \cite{Stone2010AdHA}, ZSC focuses on learning a policy that can flexibly coordinate at test time without further fine-tuning, and the pool of previously-unseen partners is not known ahead of time, i.e., it is impossible to train an agent to bias towards some known idiosyncrasies of the partner agents. On the other hand, ad-hoc teaming attempts to achieve coordination by having agents update their policy while interacting with other agents, and the the pool of partner agents can be known in advance (though it can be of arbitrarily large size, thus the knowledge can be practically impossible to exploit).

\subsection{AI for Hanabi}

\begin{table}[]
\centering
\caption{Survey of Rule-based and Learning-Based Hanabi AI Performance in 2-Player Games}
\label{tab:hanabi_ai_scores}
\small
\begin{tabular}{lllccc}
\toprule
\textbf{Hanabi AI} & \textbf{Original Author} & \textbf{References} & \textbf{Self-Play} & \textbf{Cross-Play} & \textbf{Human-Play} \\ \midrule

Van den Bergh     & Van den Bergh 2016 &  \citep{van2016aspects,walton2017evaluating,goodman2019re} & 13.8 & 10.8  & --  \\

Self-Recognition  & Osawa 2015 & \citep{osawa2015solving,goodman2019re} & 15.9 & -- & --  \\

% IGGI              & Walton-Rivers 2017 & \citep{walton2017evaluating,sarmasi2021hoad} & 17.0 & 11.0 & -- \\

Piers             & Walton-Rivers 2017 & \citep{walton2017evaluating,sarmasi2021hoad} & 17.3 & 11.2 & -- \\

Intentional AI    & Eger 2017          & \citep{eger2017intentional,canaan2020generating} & 12.6 & \textbf{13.8} & \textbf{15.0} \\

Expectimax        & Bouzy 2017 &  \citep{bouzy2017playing,goodman2019re} & 19.0 & -- & --  \\

MirrorSituational & Canaan 2018 & \citep{canaan2018evolving,goodman2019re}  & 20.1  & 12.4 & -- \\

RIS-MCTS          & Goodman 2019       & \citep{goodman2019re,canaan2020generating} & 20.6 & 13.3 & --  \\

WTFWThat          & J. Wu 2018         & \citep{wu2018hanabi,bard2020hanabi,goodman2019re} & 19.5 & -- & -- \\

FireFlower     & D. Wu 2018          & \citep{wu2018fireflower,bard2020hanabi,foerster2019bayesian} & 22.7 & -- & -- \\

Implicature AI    & Liang 2019         & \citep{liang2019implicit,canaan2020generating} & 18.9 & -- & \textbf{$\sim$ 15} \\ 

\textbf{SmartBot (SB)}     & O'Dwyer 2019          & \citep{odwyer2019hanabi,bard2020hanabi,goodman2019re} & \textbf{23.0} & -- & -- \\

\midrule

IL-Valuebot       & Sarmasi 2021       & \citep{sarmasi2021hoad,odwyer2019hanabi} & 18.0 & -- & -- \\

% IL-IGGI           & Sarmasi 2021       & \citep{sarmasi2021hoad} & 16.2 & -- & --\\

% IL-Piers          & Sarmasi 2021       & \citep{sarmasi2021hoad} & 15.9 & -- & -- \\

% IL-Rainbow        & Sarmasi 2021       & \citep{sarmasi2021hoad} & 18.1 & -- & -- \\

% Rainbow           & Bard 2020          & \citep{bard2020hanabi,hu2019simplified} & 20.6 & 1.2 & -- \\

ACHA              & Bard 2020          & \citep{bard2020hanabi,hu2019simplified} & 22.7 & 1.0 & -- \\

BAD               & Foerster 2019      & \citep{foerster2019bayesian,bard2020hanabi,hu2019simplified} & 23.9 & -- & -- \\

SAD               & Hu 2019            & \citep{hu2019simplified,hu2020other} & 24.0 & 3.0 & -- \\

SAD+AUX           & Hu 2019            & \citep{hu2019simplified,hu2020other} & 24.0 & 21.1 & 9.2 \\

SPARTA            & Lerer 2020         & \citep{lerer2020improving,canaan2020generating}  & \textbf{24.6}  & -- & -- \\

\textbf{Other-Play (OP)}   & Hu 2020            & \citep{hu2020other} & 24.1 & \textbf{22.5} & \textbf{15.8} \\ \bottomrule
\end{tabular}
\caption*{We provide Hanabi AI agents' scores as they are reported in existing literature. This gives a notional perspective of the current state of the art; however, we note that evaluation conditions have not been standardized throughout previous literature (i.e. number of game seeds for self-play, pool of teammates for cross-play, etc.). %
Therefore some caution is needed when making direct comparisons between reported scores. Please see respective literature for details on evaluation conditions.}
\end{table}

%The increased attention on Hanabi as an AI benchmarking environment has led to several research competitions \citep{walton20192018,bard2020hanabi} and an increasingly sophisticated collection of Hanabi-playing AI agents. %
Table \ref{tab:hanabi_ai_scores} summarizes the most salient published Hanabi AI agents in 2-player games with game score in self-play, cross-play, and human-play. %
The table is separated into \emph{rule-based} (top) and \emph{learning-based} (bottom) agent types. %
Rule-based agents have a policy composed of a predefined set of rules to follow given any particular game situation, and the rules are often derived from human domain knowledge. Learning-based agents, on the other hand, use statistical learning methods to adjust the parameters of their policy. The mechanism governing what action the policy will choose is learned via the agent’s experience, without the need for human domain knowledge.

Most early work focused on rule-based AI \cite{osawa2015solving,van2016aspects,walton2017evaluating}. Goodman won the 2018 CoG Hanabi Competition \citep{walton20192018} with a rule-based Monte Carlo Tree Search agent. %
O'Dwyer's SmartBot (SB) is the highest performing rule-based agent created to date, in terms of self-play~\citep{odwyer2019hanabi,bard2020hanabi,goodman2019re}. 
Due to its state-of-the-art SP score, as well as a readily available implementation~\citep{odwyer2019hanabi}, we select SB as our rule-based agent in the human-AI experiments presented in Sections \ref{sec:methods} and \ref{sec:results}. %
We found no previous work that evaluated SB in the cross-play or human-play settings.
%We choose SmartBot as our rule-based agent for the human experiments presented in Sections \ref{sec:results} for its state-of-the-art SP scores as well as the fact that the model was readily available 

Recently developed learning-based agents have demonstrated breakthrough Hanabi performance. Sarmasi et al. provide a collection of agents trained with imitation learning that nearly match the performance of the rule-based agents from which they were trained \citep{sarmasi2021hoad}. A sequence of publications \citep{foerster2019bayesian,hu2019simplified,lerer2020improving,hu2020other} offered reinforcement learning agents that each advanced the state of the art self-play and/or cross-play performance at the time of publication. This culminated with the SPARTA \cite{lerer2020improving} and Other-Play (OP) \citep{hu2020other} agents; which, respectively, represent the highest performance to date in self-play and cross-play of any agent type.

The OP agent is selected as the learning-based agent for study in our human-AI experiments described in Sections \ref{sec:methods} and \ref{sec:results}. 
We choose OP, not only for its state-of-the-art cross-play performance, but also because it uses a learning objective function specifically designed to optimize for zero-shot coordination settings. 
%OP agents are of particular note because they use a learning objective function specifically designed to optimize zero-shot coordination settings. 
Due to this optimization, we expect that humans would subjectively prefer and objectively perform better with OP agent teammates over other AI teammates  that were not designed for the zero-shot coordination setting, such is SmartBot.% in the types of zero-shot or few-shot coordination games described Section \ref{sec:methods}.

A handful of works have conducted human-play experiments; however these works are uncommon due to the significant time and effort required to conduct such experiments. 
Both \citet{eger2017intentional} and \citet{liang2019implicit} propose rule-based AI derived from Gricean maxims \citep{grice1975logic} and achieved an average human-play score of approximately 15.0 over a set of experiments that included hundreds of human participants. The learning-based Other-Play agent is arguably the highest performer in terms of human-play with an average score of 15.8, however the experiments were run with a much smaller pool of human participants \citep{hu2020other}. No prior works were found that provide a comparative  study of rule-based and learning-based Hanabi AI in human experiments.

\subsection{Human-AI Teaming and Metrics}

Human-machine interaction evaluations typically consider two broad categories of metrics: objective performance metrics (raw score, error rates, accuracy, time required, etc.), and subjective team- or human-focused metrics (situation awareness, trust, workload, etc.). Measurement of the former is heavily task-dependent, and are often the primary evaluation metrics of AI systems. The latter, however, can also involve metrics for probing systems and teams in a quantitative way, with important implications for how technology is used and adopted \citep{parasuraman1997humans}.

One key metric of teaming is \emph{trust}, which is defined by Lee and See as ``the attitude that an agent will help achieve an individual’s goals in a situation characterized by uncertainty and vulnerability'' \citep{lee2004trust}. Potential difficulties with trust include trust \emph{calibration} (whether one's trust of an agent is commensurate with its capabilities) and trust \emph{resolution} (whether the range of situations where a human trusts a system is commensurate with its range of capabilities).
%Poor trust calibration or resolution may result in \emph{overtrust} where trust exceeds system capabilities, or \emph{distrust} where system capabilities exceed trust. 
%A particularly poignant example of overtrust in automation lies in Iran Air Flight 655, a civilian airliner which was destroyed by the United States Navy guided missile cruiser  USS \emph{Vincennes} after the ship's Aegis system misidentified the airliner as an F-14 fighter, despite its flight path following one known to be used by civilian airlines \citep{blackhurst2011autonomy}. 
%Trust is typically measured via subjective ratings of related descriptors (e.g. trustworthy, deceitful, suspicious, etc.) applied to other agents.

Closely related to trust are the notions of \emph{legibility} (being expressive of one's intent) and \emph{predictability} (matching one's expectations) \citep{dragan2013legibility}. In measuring teaming, one might directly ask humans about their teammates' intent and their own expectations, particularly in relation to shared goals. Dragan et al. \citep{dragan2013legibility} argue that these two ideas trade off in the context of robot motion, but that legibility is the more important factor when working in close collaboration with a human. Similar arguments may also be made in the context of Hanabi, where maximum legibility is key to guiding a team's actions in the tightly-coupled, imperfect-information scenario.

Hoffman \cite{hoffman2019evaluating} proposes a set of subjective and objective methods (overlapping with the aforementioned ideas) to measure \emph{fluency} in human-robot interaction, defined as the ``coordinated meshing of joint activities between members of a well-synchronized team.'' Fluency encompasses subjective elements such as teamwork, trust, and positive perception; and objective measures of timing and concurrency. %, since its main application is with embodied robotics.

Much of the contemporary work on learning-based agents has not evaluated such teaming metrics, due to their focus on single-player \cite{bellemare2013arcade} or adversarial games \citep{silver2018general,vinyals2017starcraft}. In this work, we consider the use of both objective performance and subjective teaming-based measurement of human-AI pairs in the context of Hanabi. We aim to provide a more thorough analysis of the human-AI teaming data than previously-discussed Hanabi HMT studies \citep{eger2017intentional,liang2019implicit,hu2020other}. Rather than proposing a new AI with better raw performance, we are interested in the subjective aspects of human-AI teaming with state-of-the-art rule- and learning-based AI, and how they might drive future AI development.

\section{Methods}
\label{sec:methods}

\subsection{Human-AI Teaming Experiment}
\label{subsec:humanAITeamingExperiment}

Experiments consisted of two-player Hanabi games played by teams of one human participant and one AI agent. Experiments aimed to measure the objective team performance and subjective human reactions to different types of AI. The AI agents used in experiments were the Other-Play RL agent \cite{hu2020other} (specifically the \emph{OP+SAD+AUX} agent, hereafter referred to as the \emph{OP} bot) and the ``SmartBot'' agent \cite{odwyer2019hanabi} (hereafter referred to as the \emph{SB} bot), both of which have MIT Licenses. These agents were chosen because they were the top-performing learning-based and rule-based Hanabi AI, respectively at the time of the experiment.

Participants were first introduced to the experiment and the rules of Hanabi as defined in Section \ref{subsection:hanabi}. The experiment followed with a brief familiarization game so participants could acquaint themselves with the game interface. Then, each participant played two sets (blocks) of three games, with each set using a different AI teammate. The participant was not informed which AI teammate played each block of games and the order of the AI teammates was counterbalanced over the course of the study. Participants answered Likert scale surveys after each game (Table \ref{tab:post-game-and-post-experiment}, left), a NASA Task Load Index (TLX) survey after each block \citep{hart1988development}, and a Likert scale survey after both blocks that directly compared their experience with both agents (Table \ref{tab:post-game-and-post-experiment}, right). Likert scale survey questions were largely derived from a compilation of similar questions in Hoffman et al. \cite{hoffman2019evaluating}. Unless otherwise noted, Likert scales were arranged with 1 corresponding to ``strongly disagree'' and 7 corresponding to ``strongly agree.''

A total of 29 adult participants completed an experiment and each provided written, informed consent. The protocol was approved by the MIT Committee on the Use of Humans as Experimental Subjects (protocol E-2520) and the United States Department of Defense Human Research Protection Office (protocol MITL20200003).
Participants received a \$10 USD gift card at the end of their experiment, and the highest-scoring participant received an additional \$50 gift card. Each experiment took approximately 1.5 to 2.5 hours. The total amount spent on participant payment was \$350, with one participant being paid twice due to technical difficulties ending their initial session early. Experiments were conducted virtually, with all interactions occurring through video-conference, online surveys, and the Hanabi game interface, adapted from \cite{lerer2020improving}. No personally-identifiable information was collected, and no significant risk to participants was expected.

Liang et al. \citep{liang2019implicit} conducted human experiments with subjective survey questions; however the only significant result reported was that humans were more likely to mistake Implicature AI as another human.
Although Hu et al. \cite{hu2020other} conducted some human experiments with the OP bot and a self-play RL bot, their experiments only focused on game score, involved only one game with each agent, and did not consider participant expertise in Hanabi.

\begin{table*}[ht]
\centering
\caption{Post-Game (Left) and Post-Experiment (Right) Evaluation Statements}
\label{tab:post-game-and-post-experiment}
\small
\begin{tabular}{l|l}
\toprule
  & \bf{7-point Likert Scale statement}   \\ \hline
G1 & I am playing well.\\
G2 & The agent is playing poorly.\\
G3 & The team is playing well.\\
G4 & This game went well.\\
G5 & The agent and I have good teamwork. \\
& \emph{[fluency]}\\
G6 & The agent is contributing to the \\
& success of the team.\\
G7 & I understand the agent's intentions. \\
& \emph{[legibility]} \\
G8 & The agent does not understand my \\
&intentions. \emph{[legibility]}\\
G9 & I feel comfortable playing with this agent.\\
G10 & I do not trust the agent. \emph{[trust]}\\
G11 & The agent is not a reliable teammate. \\
& \emph{[predictability]}\\
G12 & I am not confident in my gameplay.\\
\end{tabular}
\begin{tabular}{l|l}
\toprule
  & \bf{7-point Likert Scale statement}\\ \hline
E1 & Which agent did you prefer playing with?                 \\
E2 & Which agent did you trust more? \emph{[trust]}                  \\
E3 & Which agent did you understand more?\\
& \emph{[legibility]}        \\
E4 & Which agent understood you better? \emph{[legibility]}          \\
E5 & Which agent was the better Hanabi player?                \\
E6 & Which agent was more reliable? \emph{[predictability]}           \\
E7 & Which agent had a better understanding of the \\
& game on average?  \\
E8 & Which agent caused you to have a greater \\
& mental workload? \emph{[mental workload]}\\
\end{tabular}
\begin{tablenotes}
\small
\item Left: Statement order was randomized when presented to participants.
\item Right: Rating 1 is ``strongly prefer first agent'' and rating 7 is ``strongly prefer second agent.''
\item Items associated with particular human factors constructs have the construct name in brackets (not shown to participants).
\end{tablenotes}
\end{table*}

\subsection{Statistical Analysis}

We evaluated both objective and subjective metrics of human-AI teaming, with the overall hypothesis that the Other-Play RL agent (OP) is preferred over- and would outperform the rule-based SmartBot (SB) agent. As an objective measure, we consider the team score during each game. For subjective measures, we consider outcomes related to perceived performance, teamwork, legibility, comfort, trust, and workload as measured by the Likert surveys. Linear mixed-effects regression models were used for both objective and subjective variables, following the recommendation of \citet{norman2010likert}. Due to space constraints, we do not present the results of the TLX here, but include one question on mental workload from the post-experiment survey (Table \ref{tab:post-game-and-post-experiment}, E8).

Both objective and subjective omnibus tests for post-game results used second-order mixed-effects models with fixed factors of (1) AI agent, (2) self-rated Hanabi experience, (3) block (the first or second set of three games), and (4) game number (first, second, or third game within a block), and a random factor of participant number. AI agent and participant number were considered categorical variables. Post-experiment surveys were evaluated separately with one-sample $t$-tests, and a Holm–Bonferroni multiple comparison correction, since these tests were looking for differences from a neutral value on the Likert scale, and there were multiple hypotheses being tested. Additionally, correlations between self-rated experience, and post-game responses were checked against game score.

\section{Results}
\label{sec:results}

We present the results of the teaming experiment. Our results indicate that despite \emph{no significant difference in objective performance} between teaming with the two agents (Section \ref{subsec:gameScore}), human sentiment results (Sections \ref{subsec:post_game_sentiments} and \ref{subsec:post_experiment_sentiments}) show a \emph{clear preference toward a rule-based agent over a learning-based agent}.

For context, we have colored most of the plots to show the self-rated Hanabi experience level of the participant with which the data are associated, and we note that our participant pool is skewed towards higher-self-rated-experience players. Self-rated Hanabi experience was based on ratings of the demographic survey statement ``I am experienced in Hanabi'' (see supplemental materials for demographic survey results).

\subsection{Game Score}
\label{subsec:gameScore}

% \begin{figure}[htp]
%     \centering
%     %\includegraphics[width=0.6\columnwidth]{figures/scores/small-scores-boxplot.jpg}
%     %\includegraphics[width=0.6\columnwidth]{figures/scores/small-scores-histogram-bins20.jpg}
%     \includegraphics[width=0.7\columnwidth]{figures/scores/small-scores-kde-histogram-bins20.jpg}
%     % \includegraphics[width=0.45\columnwidth]{figures/scores_by_game_and_block.png}
%     \caption{Game scores by agent type. No significant differences were found when teaming with either agent. (bins=20)}
%     \label{fig:u}
% \end{figure}

% The Other-Play (OP) bot mean ($\pm$ standard deviation), median, and mode game score in human-play was $14.39\pm6.97$, 15.0, and 23.0 (10 occurrences), respectively. Likewise, SmartBot (SB) scored $14.07\pm6.57$, 14.0, and [9,20] (7 occurrences each) in human-play. %for SmartBot (SB), which is in agreement with the published results in Table \ref{tab:hanabi_ai_scores} given the small sample sizes. The game score mode was 23 (10 occurrences) for OP and 9 and 20 (each 7 occurrences) for SB. 
% While the differences in these metrics are not statistically significant, it is worth noting that OP scores higher on each metric, and thus we can say that OP \emph{does not under-perform} SB in human-play. This is notable given the human sentiment results in Sections \ref{subsec:post_game_sentiments} and \ref{subsec:post_experiment_sentiments} that show a \emph{clear preference toward SB in spite of no measurable objective performance benefit of SB}.

\begin{figure}[htp]
    \centering
    \includegraphics[width=0.49\columnwidth]{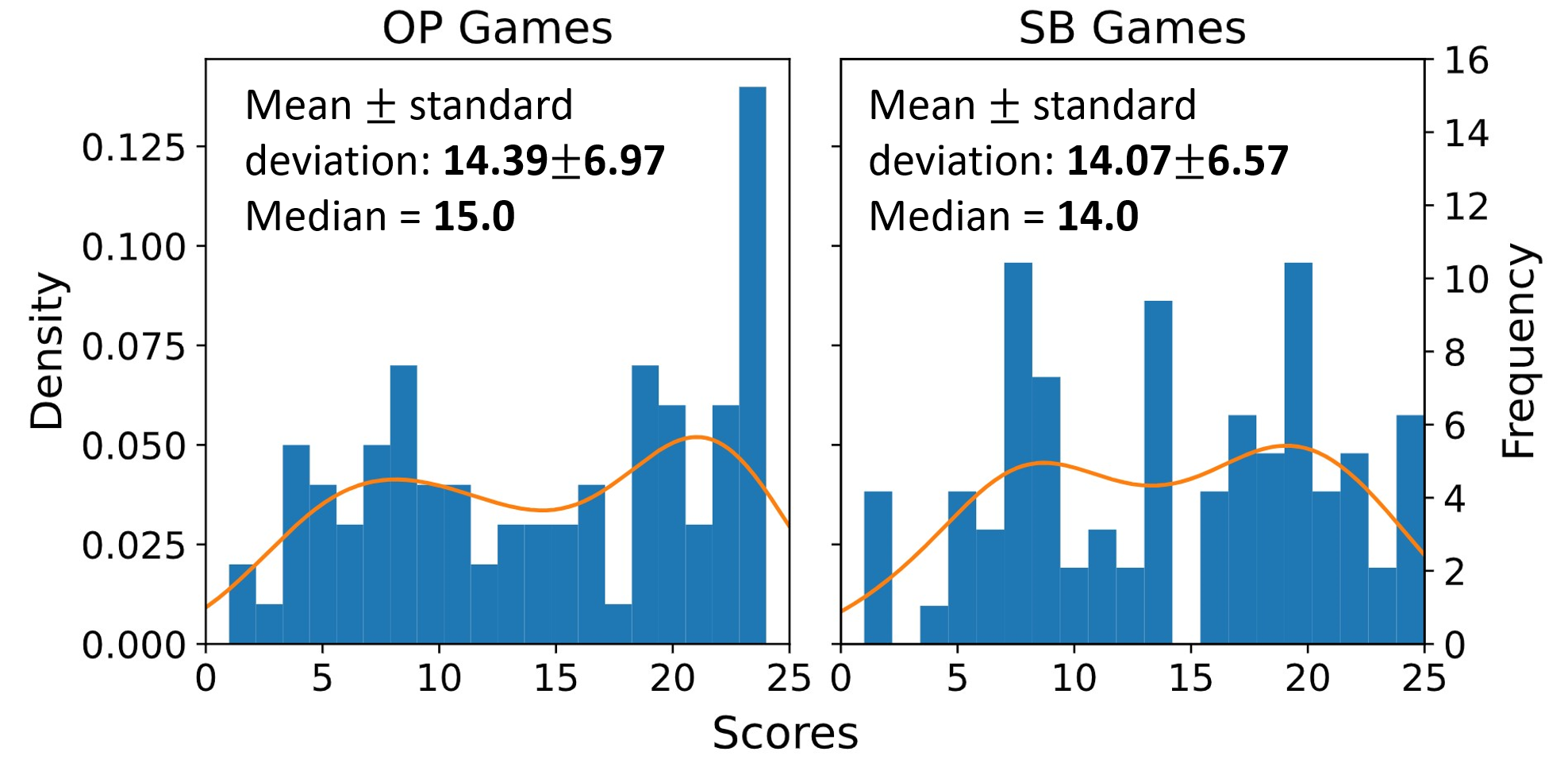} \includegraphics[width=0.49\columnwidth]{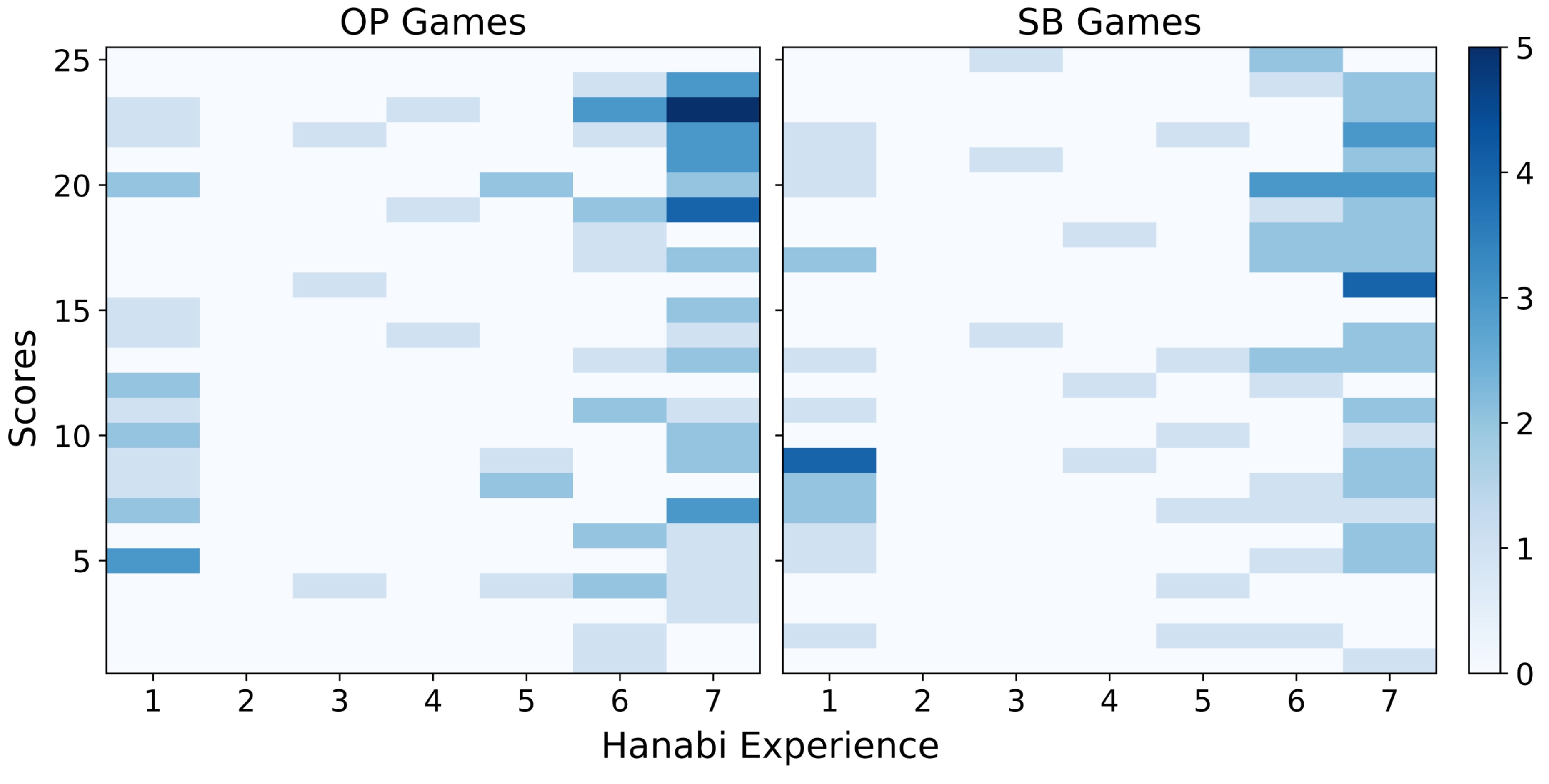}
    \caption{Game scores by agent type (left) and self-rated player experience (right). No significant differences were found when teaming with either agent, and correlation with self-rated experience was significant only for SmartBot games.}
    \label{fig:score_histogram_and_heat_map}
\end{figure}

The mixed-effects regression \emph{did not} support score differences due to agent type ($t(158)=0.374), p=0.709$) or self-rated Hanabi experience ($t(158)=1.228, p=0.221$) (Figure \ref{fig:score_histogram_and_heat_map}). This means that the data do not support OP under-performing relative to SB.

The dependent variable of game score only has a significant effect of block ($b = 5.282, t(158)=2.636, p=0.009$), and a near-significant effect of game number ($b=2.957, t(158)=1.887, p=0.061$). The positive regression slope in both cases indicates that this is likely a learning effect over time; i.e. the human participant adapts his/her play over time. 

% \begin{figure}[htp]
%     \centering
%     %\includegraphics[width=0.45\columnwidth]{figures/SB-heatmap-scores.png}
%     %\includegraphics[width=0.45\columnwidth]{figures/OP-heatmap-scores.png}
%     \includegraphics[width=0.65\columnwidth]{figures/scores/small-heatmap-scores-merged.jpg}
%     \caption{Heat maps of score vs self-rated Hanabi experience, divided by agent type. Correlation was significant only for SmartBot games.}
%     \label{fig:score_experience_corr}
% \end{figure}

Since score is the primary performance metric of interest in Hanabi, and acts as the reward function for RL agents in this domain, we examined a few additional correlations with score. Note that since these are bivariate correlations, they are not as complete as the aforementioned statistical regression.

Self-rated Hanabi experience and score had a small, but significant positive correlation when pooling both agents' games ($p=0.0053$, $r=0.214$). Correlation remains significant for the subset of SB games ($p=0.0023$, $r=0.247$), but is not significant for OP games ($p=0.0867$, $r=0.1881$), indicating that for this bivariate analysis, experience only correlates with score for SB, not with OP.

A small, but significant positive correlation was found between subjective team performance (G3, G4) and score ($p=0.0003$, $r=0.275$ and $p=0.0002$, $r=0.280$). However, subjective ratings of self- and agent-performance (G1, G2) were not significantly correlated to score.

When participants were ordered by the absolute differences in their total scores with each agent, the ten subjects with the greatest differences (max/min point differences of 57 and 17), six of the ten had better performance with OP, but when considering only the top five subjects in terms of score difference (max/min differences of 57 and 28), only the participant with the greatest difference performed better with SB, and all four others performed better with OP. A full plot of scores by subject and agent type is in the supplemental materials.

\subsection{Post-Game Sentiments}
\label{subsec:post_game_sentiments}

Significant effects in post-game subjective measures are summarized in Table \ref{tab:post-game-stats}. Although the statistical model considers factors of block and game, which were sometimes significant, we do not report these main effects or their interactions in the table, in order to focus on the independent variables of greater interest.

\begin{figure}[htp]
    \centering
    \includegraphics[width=\columnwidth]{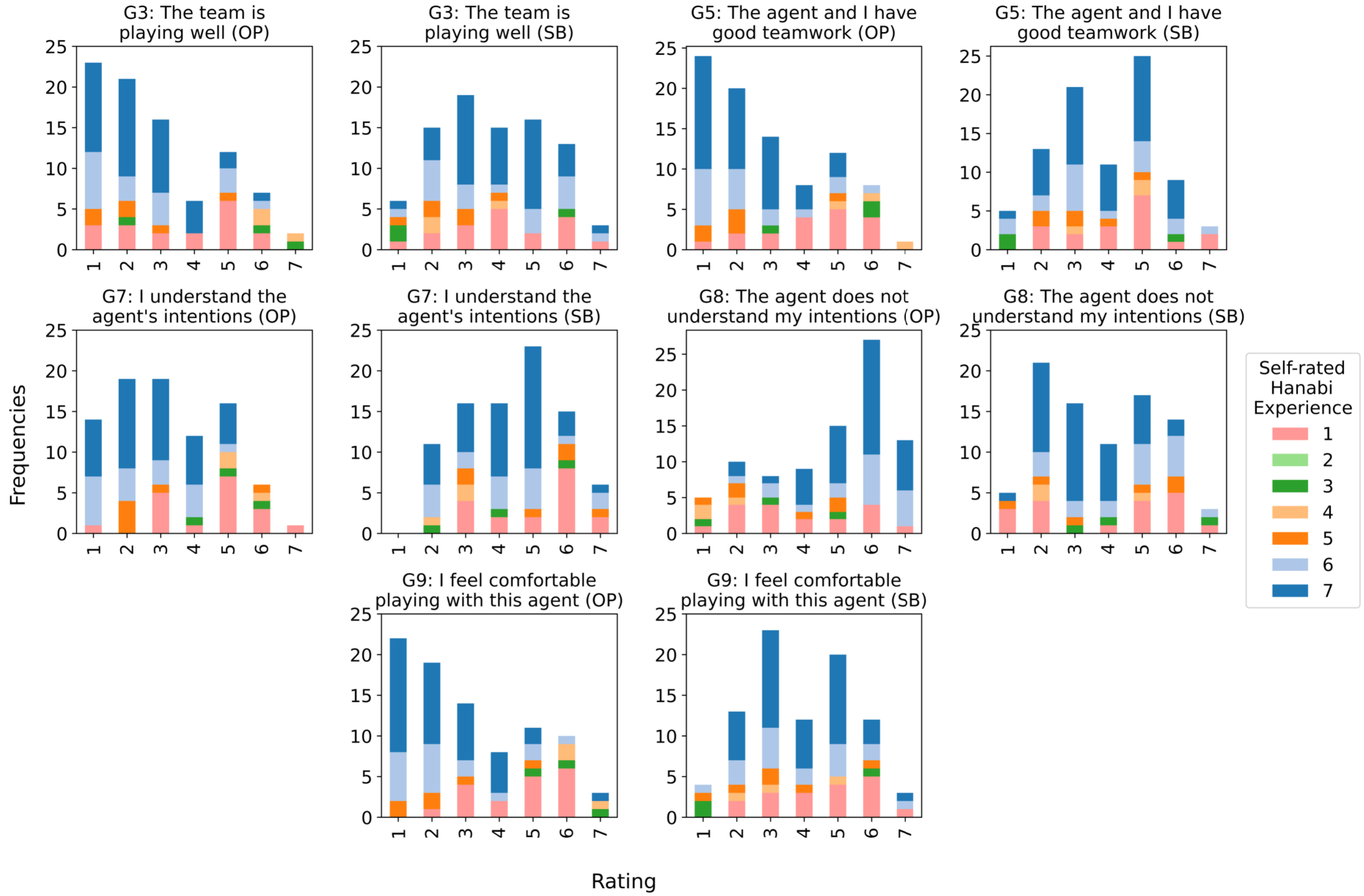}
    \caption{Participant rating for post-game questions by self-rated Hanabi experience (SB vs OP), where statistically significant differences related to factors of agent and/or experience were found (Table \ref{tab:post-game-stats}). The scale ranges from 1-7, corresponding to "strongly disagree" to "strongly agree".}
    \label{fig:post_game_questions_separate}
\end{figure}

\begin{figure}[htp]
    \centering
    \includegraphics[width=0.8\columnwidth]{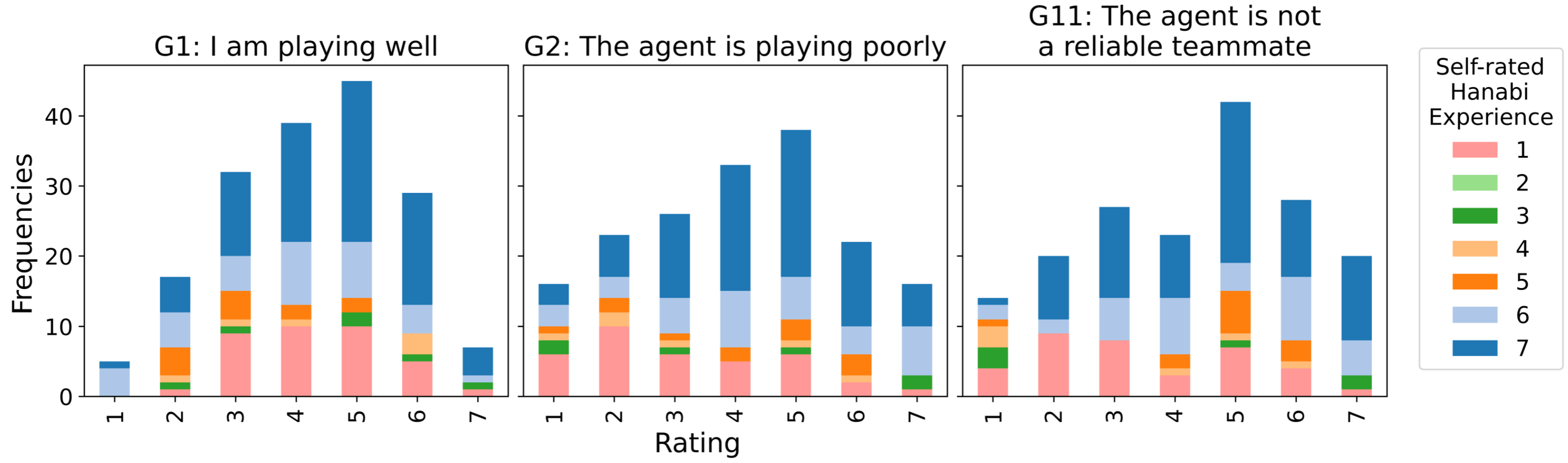}
    \caption{Participant rating for post-game questions by self-rated Hanabi experience where no statistically significant differences related to factors of agent and/or experience were found. The scale ranges from 1-7, corresponding to "strongly disagree" to "strongly agree".}
    \label{fig:post_game_questions_joint}
\end{figure}

In all cases where the interaction of agent type and Hanabi experience was found to be significant, more experienced Hanabi players rated the Other-Play agent much more negatively than the SmartBot agent, while novices rated the two similarly (Figure \ref{fig:post_game_questions_separate}). However, there was no significant difference between novice and expert ratings of the SmartBot agent. Cases where neither agent nor experience were significant factors are shown in Figure \ref{fig:post_game_questions_joint}.

\begin{table*}[ht]
\centering
\caption{Post-Game Sentiment Statistics (Statistically-Significant Factors Only)}
\label{tab:post-game-stats}
\small
\begin{tabular}{l|l|l|l}
\toprule
\bf{Dependent Variable} & \bf{Factor} & \bf{$t$} & \bf{$p$} \\ \hline
I am playing well (G1)                            & experience                & $2.698$ & $0.008$  \\ \hline
The agent is playing poorly (G2)                  & experience                & $4.044$ & $<0.0001$ \\ \hline
The team is playing well (G3)                     & agent:experience          & $-2.082$ & $0.039$ \\ \hline
The agent and I have good teamwork (G5)           & agent                     & $2.578$ & $0.011$  \\
                                                    & agent:experience          & $-3.021$ & $0.003$ \\ \hline
I understand the agent's intentions (G7)          & agent:experience          & $-2.273$ & $0.024$ \\ \hline
The agent does not understand my intentions (G8)  & experience                & $2.098$ & $0.037$  \\
                                                    & agent:experience          & $-3.166$ & $0.002$ \\ \hline
I feel comfortable playing with this agent (G9)   & agent                     & $3.302$ & $0.001$  \\ 
                                                    & agent:experience          & $-3.561$ & $<0.0001$ \\ \hline

The agent is not a reliable teammate (G11)        & experience                & $3.159$ & $0.002$  \\ \hline

\end{tabular}
\begin{tablenotes}
  \small
  \item Degrees of freedom for $t$-tests are 164 in all cases. \emph{Agent} is agent type (SmartBot or Other-Play), and \emph{experience} is self-rated Hanabi experience. \emph{Agent:experience} is the interaction effect of agent and experience
\end{tablenotes}
\end{table*}

To specifically examine the effect of self-rated player experience, we performed post-hoc pairwise comparisons in cases where experience was significant. Participants were pooled into ``novice'' ($n=10$, self-rated experience of $\leq 5$) and ``expert'' ($n=19$, self-rated experience of $>5$) groups, and comparisons were made on ratings of each agent in the cases where an interaction effect was significant (G3, G5, G7, G8, G9). The groups did not rate SB significantly differently, but experts always rated OP worse than novices did. The difference in G3 ``the team is playing well'' ($t(85)=3.551$, $p<0.001$, effect size $d=0.752$) was not as stark as the others ($t(85)=5.068$ to $5.855$, $p<0.0001$, $|d|>1.0$), but all were still clearly significant, and all but one case had large effect sizes.

\subsection{Post-Experiment Sentiments}
\label{subsec:post_experiment_sentiments}

\begin{figure}[htp]
     \centering
    \includegraphics[width=\columnwidth]{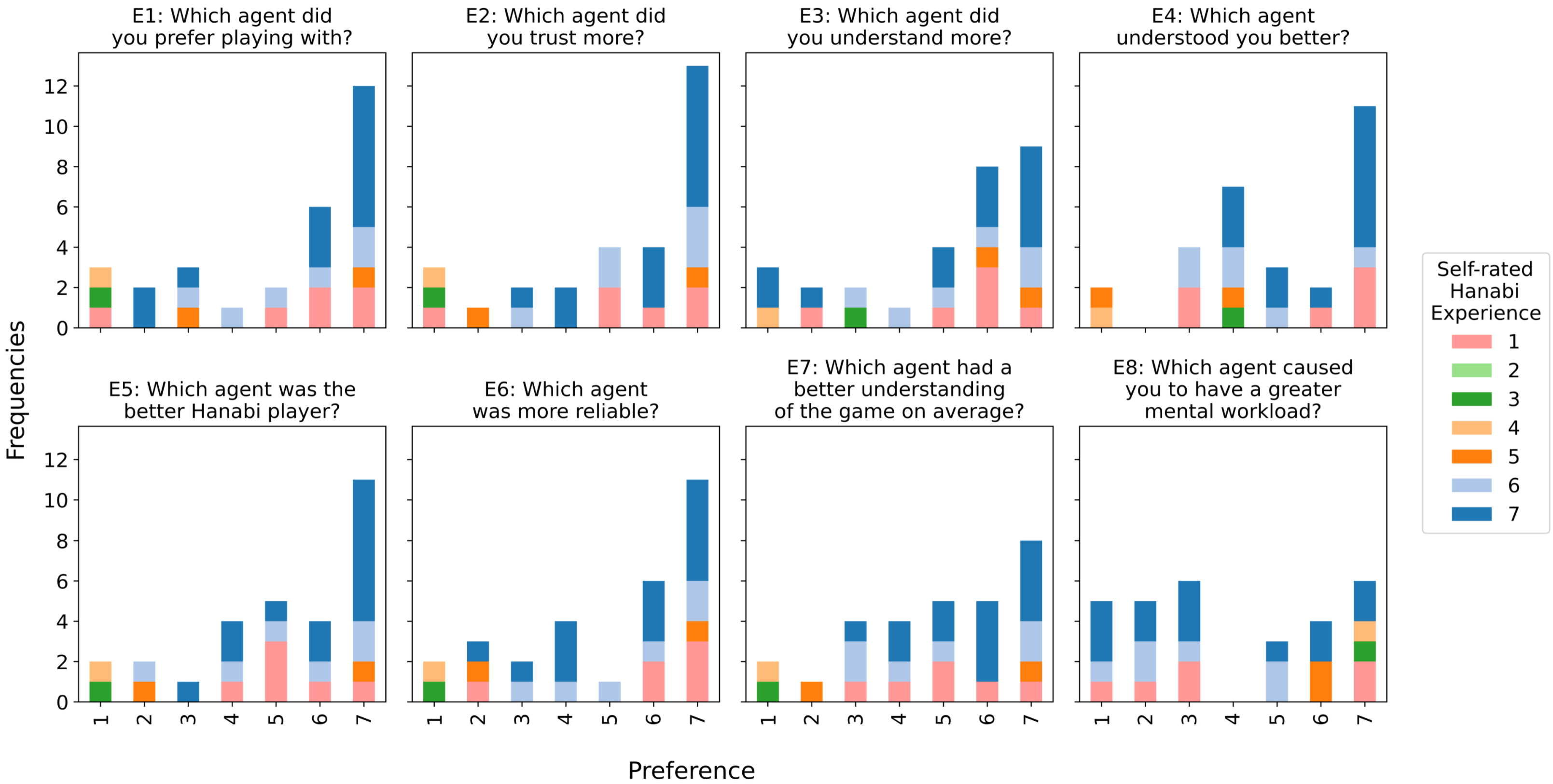}
     \caption{Post-experiment questions. All showed a statistically significant preference for the rule-based SmartBot ($p<0.05$), except the question on workload (E8). The scale ranges from 1 ("strongly prefer OP") to 7 ("strongly prefer SB").}
     \label{fig:post_experiment_all_questions}
\end{figure}

Participants' direct comparisons of the agents (Table \ref{tab:post-game-and-post-experiment}) are shown in Figure \ref{fig:post_experiment_all_questions}. The statements were presented to participants as comparing the ``first'' and ``second'' agents (without reference to agent type, which was not disclosed to participants). For ease of interpretation, we matched the responses to the agent type, and flipped the scale as appropriate. All positively-framed questions (E1 to E8) showed a strong preference to SB over OP (corrected $p<0.05$), while mental workload (E8) was split. %All reported $p$ values are with the Holm's correction.% (which results in several repeated $p$ values).

We note there were three participants who obtained a score of 24 with OP, with one of these participants obtaining 24 twice (no player achieved a 25 with OP). All three replied at the extreme end of our Likert scale (i.e., 1 or 7) for question E1 with a preference for SB. Interestingly, their cumulative scores for OP and SB, respectively, were: Participant 6 (played with OP first, self-rated experience of 7): 57 and 28; Participant 19 (SB first, experience 7): 68 and 48; Participant 20 (OP first, experience 6): 70 and 35. The participant with the highest cumulative score (Participant 2, OP first, experience 7) had cumulative scores of 68 (OP) and 54 (SB) and preferred SB with a Likert rating of 6. All participant scores are provided in the Appendix (Figure \ref{fig:participant_scores}).

% Subject commentary showed that high mental workload was often due to trying to understand the agent, or the length of the game, but low mental workload when working with OP was often caused by becoming frustrated with the agent and giving up on teaming with it. 
Participant commentary indicated that low mental workload when working with OP was often caused by frustration with the agent and giving up on teaming with it.
For example, after the OP bot failed to act on several hints from the human (\emph{``I gave him information and he just throws it away''}) a participant commented that \emph{``At this point, I don't know what the point is,''} regarding working with the agent.

\section{Discussion}
\label{sec:discussion}

\subsection{Other-Play Agent Performance}

Other-Play (OP) training is designed to avoid the creation of ``secretive'' conventions that can result from self-play training. 
However, even though OP agents can pair well with previously-unseen teammates, there are still assumptions placed on the types of teammates OP agents will encounter. %
% Most notably, OP assumes that teammates are also optimized under a zero-shot coordination setting; in other words, OP-trained agents are optimized for test-time teaming with other OP-trained agents, even if they did not encounter those other OP-agents during training \cite{hu2020other}. %
Notably, OP assumes that teammates are also optimized for zero-shot coordination \cite{hu2020other}. %
In contrast, human Hanabi players typically do not learn with this assumption.
% Although in-game communication is strictly limited to hints and action observations, pre-game convention-setting and post-game reviews are common practices for human Hanabi players..
% Thus, human learning of Hanabi may be better described as \emph{one-shot} or \emph{few-shot} coordination and does not satisfy the zero-shot assumption of OP.
Pre-game convention-setting and post-game reviews are common practices for human Hanabi players, making human learning more akin to \emph{few-shot} coordination. Furthermore, Hu et al. note that, due to practical implementation details, OP does not always avoid secretive conventions. For example, some OP agents use color hints to indicate discarding of the 5th card \citep{hu2020other}. Such conventions are not legible to a human teammate without extensive observation.

\subsection{Implications for Human-AI Teaming}
\label{subsec:implicationsForHumanAITeaming}

The results shown here have important implications for the development of learning-based, human-teaming agents. Regardless of the objective performance of AI systems, ``teaming intelligence'' is ultimately required for real-world deployment. The difference between rule-based and learning-based systems is particularly poignant here, where humans strongly favored a rule-based agent over an RL agent in many subjective metrics (and never significantly favored the RL agent) despite achieving objective scores that were not significantly different across the two agent types. Caution should be exercised when generalizing our results beyond Hanabi, though there is some indication of cross-domain consistency. The negative post-experiment sentiments expressed towards OP by some participants are similar to those expressed by participants during the DotA2 OpenaAI Five games where RL players exhibited playstyles that were not immediately understood by human players, even though they were successful \citep{OpenAI_2019, OpenAI_five_2019}.

An additional consideration is the user base of future learning-based systems. We note that between post-game and post-experiment surveys, nearly all statements/questions associated with human factors constructs (except workload) had differences related to the agent and/or the participant's experience level. Experienced players rated the OP agent particularly negatively (Figure \ref{fig:post_game_questions_separate}, Table \ref{tab:post-game-stats}), despite similarity in scores. As domain experts are likely to be the first users of AI technology in operational settings, their perception of AI is a key factor to its adoption.
Part of the experts' negativity toward OP may come from the way human experts make decisions in uncertain situations. Klein's recognition-primed decision-making model \citep{klein1993recognition} indicates that experts typically rely on a base of knowledge to recognize ``prototypical'' situations and alleviate much of the mental burden of decision-making. Working with an unpredictable and illegible teammate breaks experts' ability to rely on much of their knowledge base. This notion is also supported by the score/experience correlations (Figure \ref{fig:score_histogram_and_heat_map}) which were only significant for SB games. %Not only does the team dynamic not fit their mental model, but the team behavior now creates game states (of both explicit and implicit information) that similarly do not fit into preconceived notions of the task.

% Apart from imposing system constraints, the primary way in which human designers affect the behavior of RL agents is through reward-shaping, meaning that ``surprising'' agent behaviors are fairly common in RL, and relatively rare in expert systems \hl{citation needed? Move 37, RL superstitions, finding edge cases, etc}. Such behaviors are often treated as academic curiosities, or are not of any particular consequence in adversarial environments as long as the agent ``wins.'' However, we see that in tightly-coupled human-AI teams, such surprises can quickly lead to loss of trust, discouragement, and a desire to stop working with the AI agent on the part of the human. Much like how robotics uses safety constraints for any real-world ML training, or how ethical constraints might be imposed for some real-world deployments of machine learning systems, this work points to the potential need for ``sanity-check'' or ``trust guardrails'' to prevent AI agents from breaking the trust of their human teammates, to ensure the continued functionality of the team.

This work indicates a need for RL methods that produce legible and predictable policies.
Methods in this direction generally involve either building inherently transparent models, or procedures for post-hoc analysis. The former includes human-language-driven policy specification \cite{tambwekar2021interpretable}, or explanation learning through construction of causal diagrams \cite{Madumal2020ExplainableRL}. The latter includes saliency maps for visual domains, and action preference explanation through reward decomposition \cite{Juozapaitis2019ExplainableRL}. %There may be concern that imposing such legibility/predictability constraints may introduce a performance-transparency trade-off, but that has yet to be consistently shown. Regardless, such a trade-off may be worthwhile to ensure that a given RL agent is used rather than abandoned.

% Our subject pool is likely less biased than Liang's, since they recruited from Reddit's r/boardgames and r/cardgames

\subsection{Limitations and Future Work}
\label{subsec:limitationsAndFutureWork}

The tight coupling of coordination and performance, and the codification of communication, make Hanabi a unique test bed. Games without these properties may elicit different responses from human-AI teams, so generalization of this experiment beyond Hanabi must be treated carefully.

Due to the manual and time-intensive experimental procedure \textemdash as well as the lack of readily available implementations for many pre-existing Hanabi agents such as Intentional \citep{eger2017intentional} and Implicature AI \citep{liang2019implicit}\textemdash we were limited in our pool of participants and agents. Future work could increase both by using an online, operator-free game-play and survey platform.

While we made an effort to recruit participant with a diverse level of experience, the participant pool was still skewed towards those with higher self-rating (though it is likely less biased than the pool from \cite{liang2019implicit}, who recruited exclusively from online gaming communities). Since Hanabi performance is heavily dependent on the \emph{team}, it is difficult to obtain an objective single-player skill rating. Still, the relatively tight clustering of many (self-rated) experts' responses to OP in particular is notable.

%Another limitation is minor errors in the SmartBot's programming that caused a few instances of "obviously" bad plays (e.g. discarding a card it knows to be playable), which, according subject feedback, have particularly detrimental effect on a subject's perception of the agent. Ideally, such plays would only be found in learning-based agents.

Game outcomes and the feasibility of perfect games (i.e. score 25) are dependent on the starting deck. We did not control for deck configuration to avoid the issue encountered in Hu et al. \cite{hu2020other}, who explored a relatively narrow portion of the game space due to using only two deck seeds. %This limitation in our experiment may be overcome by the number of games involved, but it does not affect the player experience on the level of individual games.

% \subsection{Future Work}

Interesting extensions to this Hanabi experiment include varying participants' knowledge of their teammates, increasing the number of games played, and adding additional metrics of teaming, such as positive listening and positive signalling \citep{jaques2019social,lowe2019pitfalls}. Modifications to the agents may include adding logic filters to prevent learning-based agents from making ``obvious'' mistakes, or modifying the reward function to more heavily penalize ``trust-breaking'' moves. %Communication between games might also be considered, which would require more significant modification of the agent, but would mirror the kind of planning behavior that is common among human teams.
Beyond the domain of Hanabi, other candidates for human-AI teaming experiments include StarCraft (which to date has only shown adversarial RL agent performance with humans \citep{vinyals2019grandmaster}), Overcooked \citep{carroll2019utility}, and mixed cooperative-competitive environments like Bridge \citep{Yeh2018AutomaticBB} and Diplomacy~\citep{gray2020human}. %Mixed human-AI teams in the form of N vs N games might be considered. Unlike Hanabi, communication is optional in these settings, and it is possible for one player to dominate a team, so appropriate restrictions on the game, or on AI agent capabilities may be required for meaningful results. 

\section{Conclusion}
\label{sec:conclusion}

This study measured the game performance and human reactions in mixed human-AI teams in the cooperative card game Hanabi, comparing outcomes when humans were paired with a rule-based agent and when paired with a reinforcement-learning-based ``Other-Play'' agent designed to maximize zero-shot coordination. Despite achieving similar scores between these teams, human players strongly preferred working with the rule-based agent, and view the Other-Play agent quite negatively, citing their bilateral understanding, trust, comfort, and perceived performance as reasons. The ability of AI agents to team with humans is an important determinant of whether they can be deployed in many real-world situations. These results show that even state-of-the-art RL agents largely fail to convince humans that they are good teammates, and suggest that human perception of AI needs greater consideration in future AI design and development if it is to achieve real-world adoption. %for outcomes other than performance is required for what it takes to develop such agents in a way that is accepted by humans.

\newpage % acknowledgements, references, checklists, and supplemental materials do not count towards the 10-page content limit

\begin{ack}

DISTRIBUTION STATEMENT A. Approved for public release. Distribution is unlimited.
This material is based upon work supported by the Under Secretary of Defense for Research and Engineering under Air Force Contract No. FA8702-15-D-0001. Any opinions, findings, conclusions or recommendations expressed in this material are those of the author(s) and do not necessarily reflect the views of the Under Secretary of Defense for Research and Engineering.
© 2021 Massachusetts Institute of Technology.
Delivered to the U.S. Government with Unlimited Rights, as defined in DFARS Part 252.227-7013 or 7014 (Feb 2014). Notwithstanding any copyright notice, U.S. Government rights in this work are defined by DFARS 252.227-7013 or DFARS 252.227-7014 as detailed above. Use of this work other than as specifically authorized by the U.S. Government may violate any copyrights that exist in this work.

The authors would like to thank our experiment participants for their time. %
We thank Hengyuan Hu for providing the Other-Play model. % 
We thank Peter Morales for his guidance during the early phase of this work.

% Use unnumbered first level headings for the acknowledgments. All acknowledgments
% More information about this disclosure can be found at: \url{https://neurips.cc/Conferences/2021/PaperInformation/FundingDisclosure}.

% Do {\bf not} include this section in the anonymized submission, only in the final paper. You can use the \texttt{ack} environment provided in the style file to autmoatically hide this section in the anonymized submission.
\end{ack}

\medskip

\bibliographystyle{plainnat}

\bibliography{references.bib}

%%%%%%%%%%%%%%%%%%%%%%%%%%%%%%%%%%%%%%%%%%%%%%%%%%%%%%%%%%%%
\section*{Checklist}

%%% BEGIN INSTRUCTIONS %%%
% The checklist follows the references.  Please
% read the checklist guidelines carefully for information on how to answer these
% questions.  For each question, change the default \answerTODO{} to \answerYes{},
% \answerNo{}, or \answerNA{}.  You are strongly encouraged to include a {\bf
% justification to your answer}, either by referencing the appropriate section of
% your paper or providing a brief inline description.  For example:
% \begin{itemize}
%   \item Did you include the license to the code and datasets? \answerYes{See Section~\ref{gen_inst}.}
%   \item Did you include the license to the code and datasets? \answerNo{The code and the data are proprietary.}
%   \item Did you include the license to the code and datasets? \answerNA{}
% \end{itemize}
% Please do not modify the questions and only use the provided macros for your
% answers.  Note that the Checklist section does not count towards the page
% limit.  In your paper, please delete this instructions block and only keep the
% Checklist section heading above along with the questions/answers below.
%%% END INSTRUCTIONS %%%

\begin{enumerate}

\item For all authors...
\begin{enumerate}
  \item Do the main claims made in the abstract and introduction accurately reflect the paper's contributions and scope?
    \answerYes{}
  \item Did you describe the limitations of your work?
    \answerYes{See Section \ref{subsec:limitationsAndFutureWork}}
  \item Did you discuss any potential negative societal impacts of your work?
    \answerYes{See Section \ref{subsec:implicationsForHumanAITeaming}}
  \item Have you read the ethics review guidelines and ensured that your paper conforms to them?
    \answerYes{}
\end{enumerate}

\item If you are including theoretical results...
\begin{enumerate}
  \item Did you state the full set of assumptions of all theoretical results?
    \answerNA{}
	\item Did you include complete proofs of all theoretical results?
    \answerNA{}
\end{enumerate}

\item If you ran experiments...
\begin{enumerate}
  \item Did you include the code, data, and instructions needed to reproduce the main experimental results (either in the supplemental material or as a URL)?
    \answerYes{See supplemental materials.}
  \item Did you specify all the training details (e.g., data splits, hyperparameters, how they were chosen)?
    \answerNA{We did not train new models.}
	\item Did you report error bars (e.g., with respect to the random seed after running experiments multiple times)?
    \answerYes{We reported dispersion measures and test statistics where appropriate. All histograms represent exact counts of the data.}
	\item Did you include the total amount of compute and the type of resources used (e.g., type of GPUs, internal cluster, or cloud provider)?
    \answerNo{Compute was not significant since we were not training new models.}
\end{enumerate}

\item If you are using existing assets (e.g., code, data, models) or curating/releasing new assets...
\begin{enumerate}
  \item If your work uses existing assets, did you cite the creators?
    \answerYes{See Section \ref{subsec:humanAITeamingExperiment}}
  \item Did you mention the license of the assets?
    \answerYes{See Section \ref{subsec:humanAITeamingExperiment}}
  \item Did you include any new assets either in the supplemental material or as a URL?
    \answerNA{}
  \item Did you discuss whether and how consent was obtained from people whose data you're using/curating?
    \answerYes{See Section \ref{subsec:humanAITeamingExperiment}}
  \item Did you discuss whether the data you are using/curating contains personally identifiable information or offensive content?
    \answerYes{See Section \ref{subsec:humanAITeamingExperiment}. No PII was collected, and no offensive content was used/shown.}
\end{enumerate}

\item If you used crowdsourcing or conducted research with human subjects...
\begin{enumerate}
  \item Did you include the full text of instructions given to participants and screenshots, if applicable?
    \answerYes{See supplemental materials.}
  \item Did you describe any potential participant risks, with links to Institutional Review Board (IRB) approvals, if applicable?
    \answerYes{See Section \ref{subsec:humanAITeamingExperiment}. No significant participant risks were expected, no adverse events occurred during the course of the experiment. IRB approval link and institution name removed for anonymous review stage; will be added back for for later stage.}
  \item Did you include the estimated hourly wage paid to participants and the total amount spent on participant compensation?
    \answerYes{See Section \ref{subsec:humanAITeamingExperiment}}
\end{enumerate}

\end{enumerate}

%%%%%%%%%%%%%%%%%%%%%%%%%%%%%%%%%%%%%%%%%%%%%%%%%%%%%%%%%%%%

% \appendix
\newpage
\section*{Supplemental Materials}
% Optionally include extra information (complete proofs, additional experiments and plots) in the appendix.
% This section will often be part of the supplemental material.

For data transparency and completeness, we detail our participant recruitment process, participant instructions, relevant aggregate data (demographics, scores, surveys), as well as the results from statistical tests that we conducted, but were not part of the main paper due to space constraints. We do not include the results of the NASA Task Load Index survey here because those were not analyzed for this study.

\subsection{Participant Recruitment}

Participants for this experiment were recruited via convenience and snowball sampling, with initial emails to MIT research groups and social mailing lists, as well as some for other Cambridge-areas groups. We note that 6 out of 29 participants belonged to the \emph{hanab.live} Hanabi gaming community. Other than those, participants were novices to Hanabi, did not play consistently, or came from several distinct and unrelated Hanabi groups.

\subsection{Introductory Slides and Game Interface}

These are the slides shown to experiment participants at the very beginning of the session. All subjects were shown the same slides.

\begin{figure}[H]
    \centering
    \includegraphics[width=12cm]{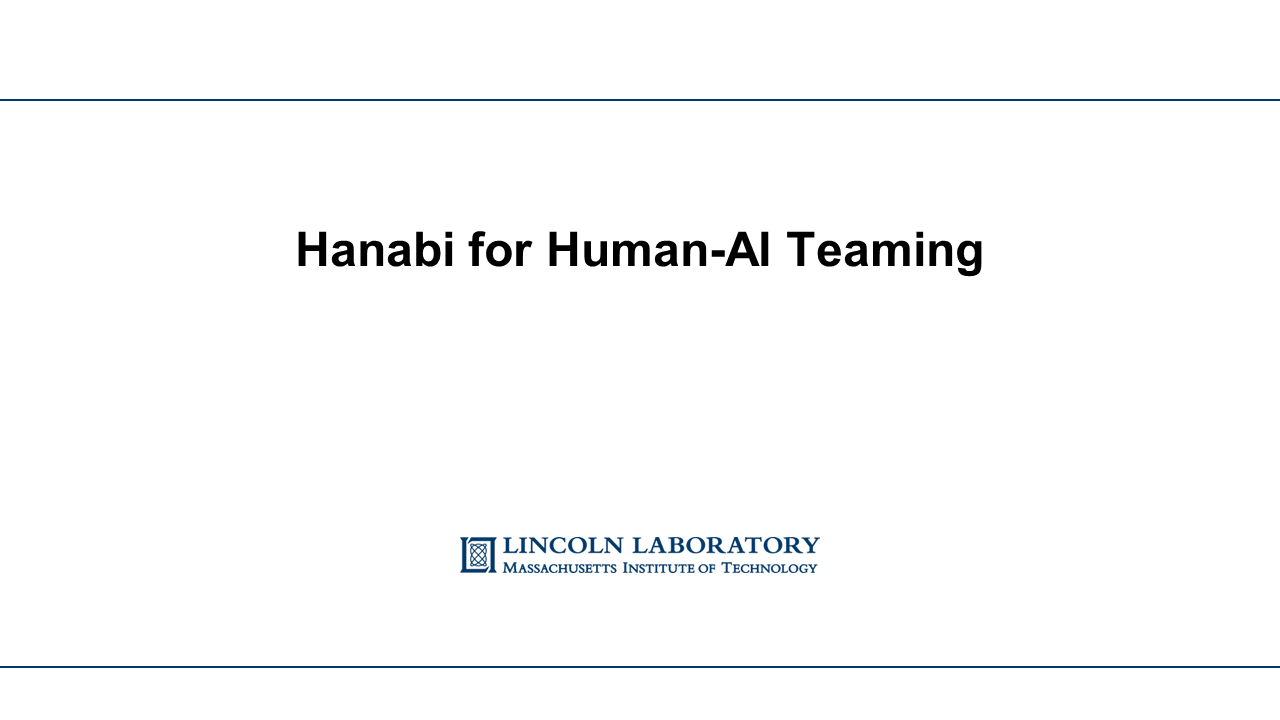}
\end{figure}
\begin{figure}[H]
    \centering
    \includegraphics[width=12cm]{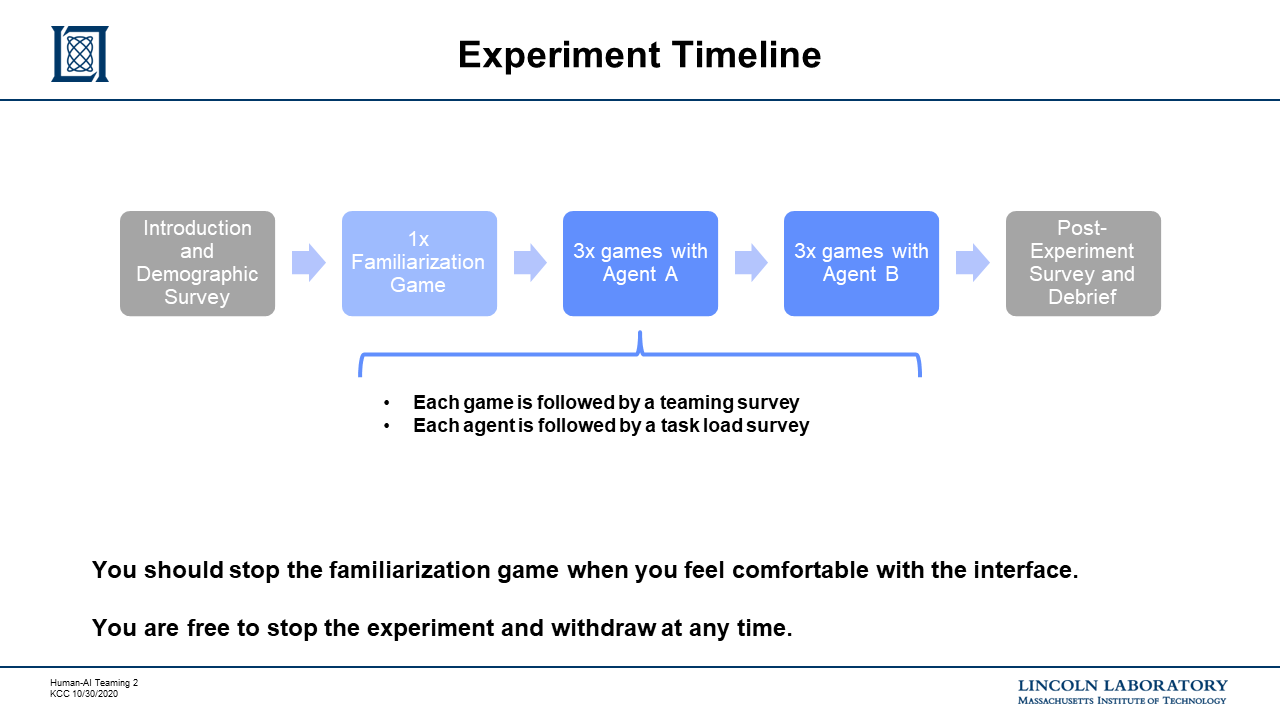}
\end{figure}
\begin{figure}[H]
    \centering
    \includegraphics[width=12cm]{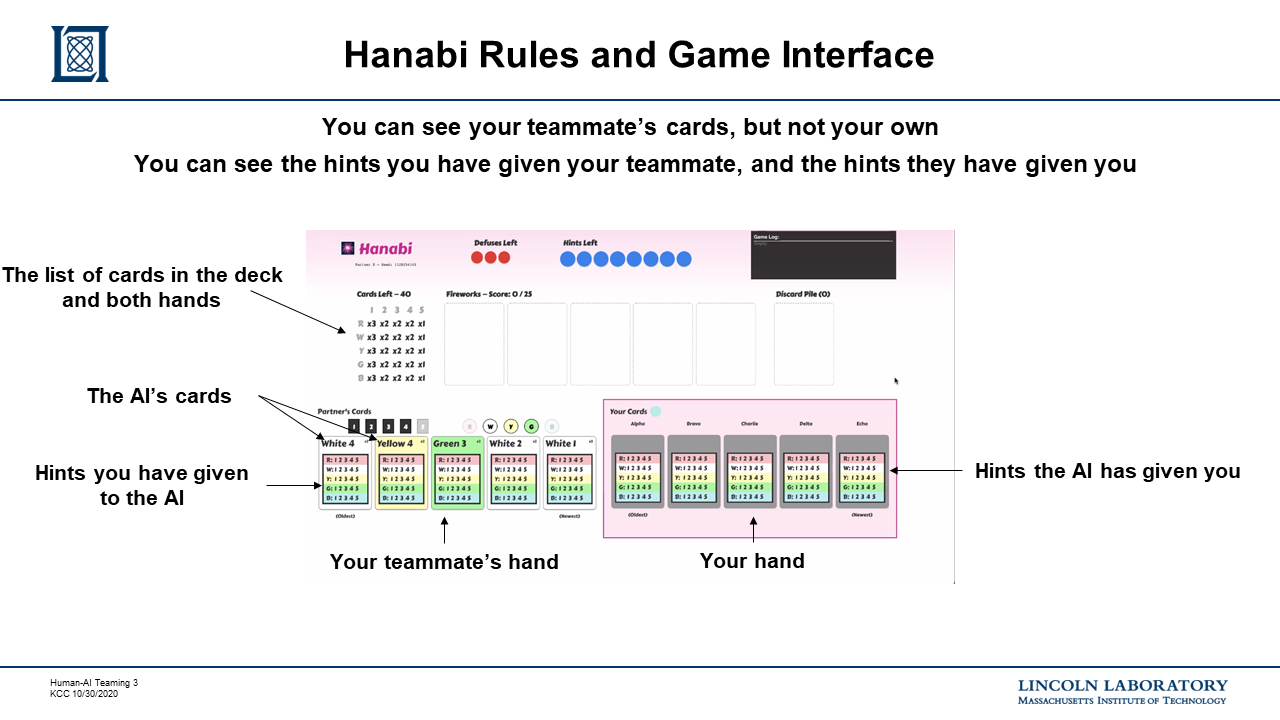}
\end{figure}
\begin{figure}[H]
    \centering
    \includegraphics[width=12cm]{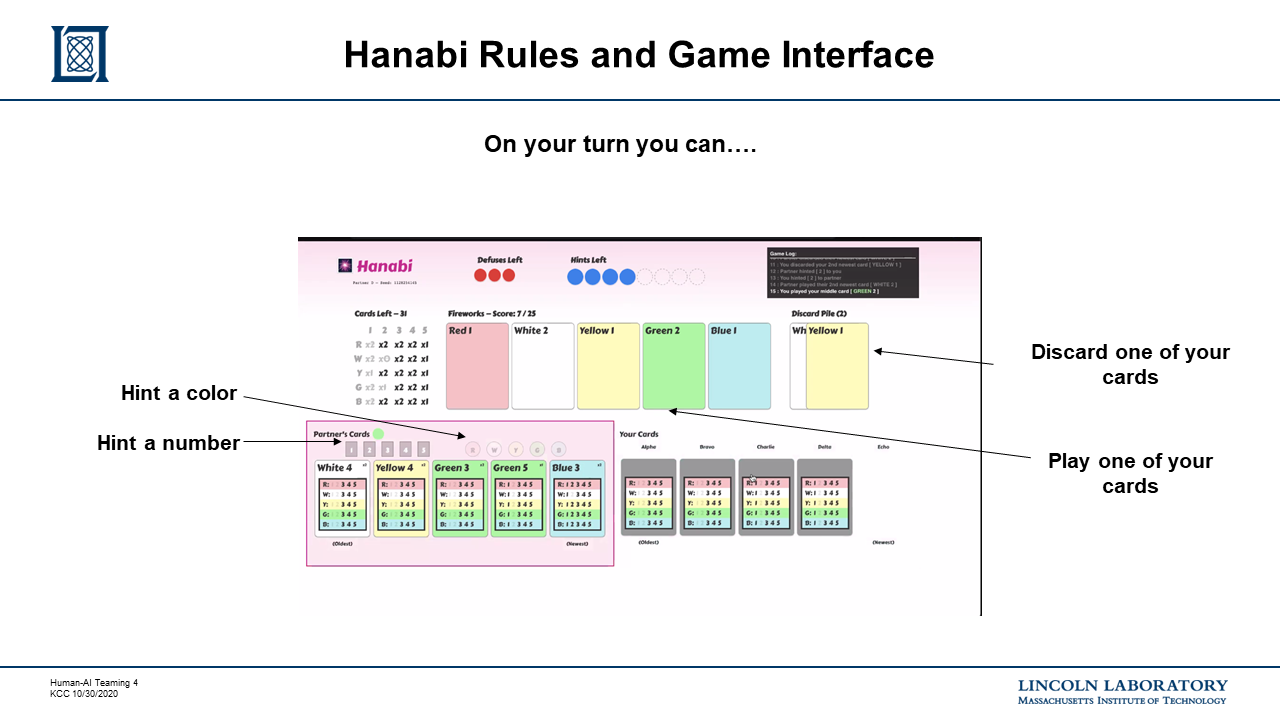}
\end{figure}
\begin{figure}[H]
    \centering
    \includegraphics[width=12cm]{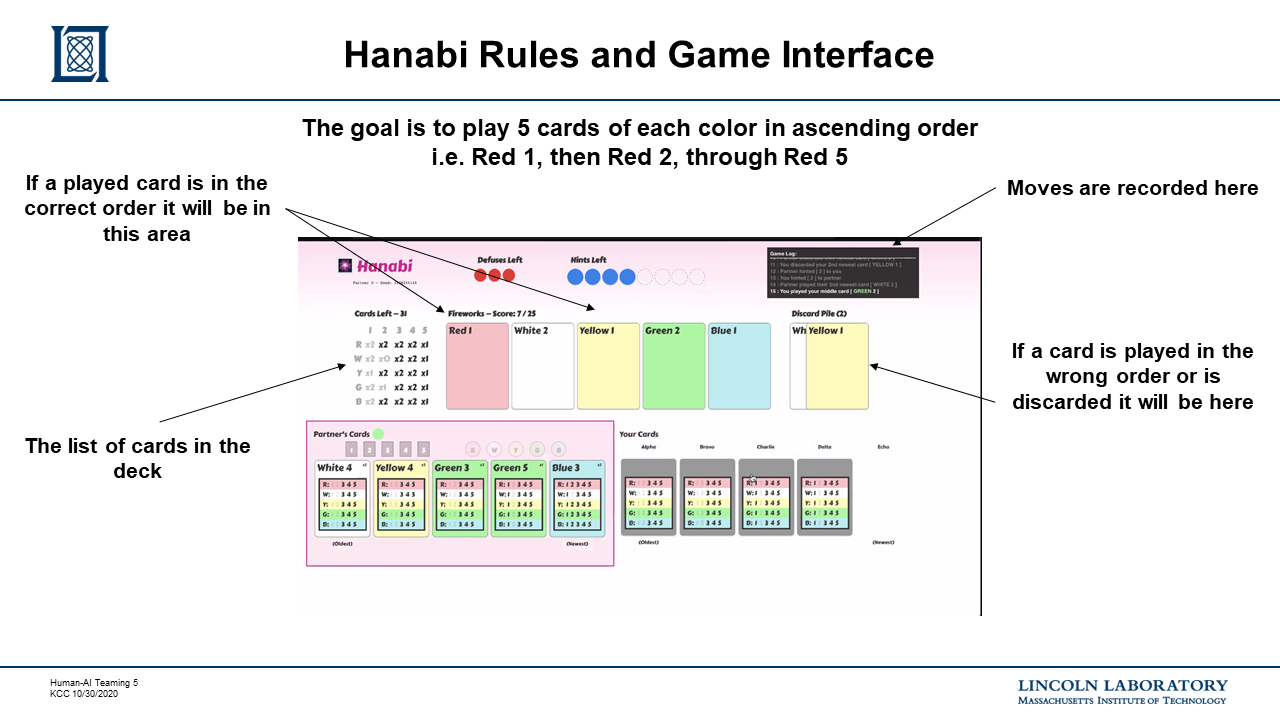}
\end{figure}
\begin{figure}[H]
    \centering
    \includegraphics[width=12cm]{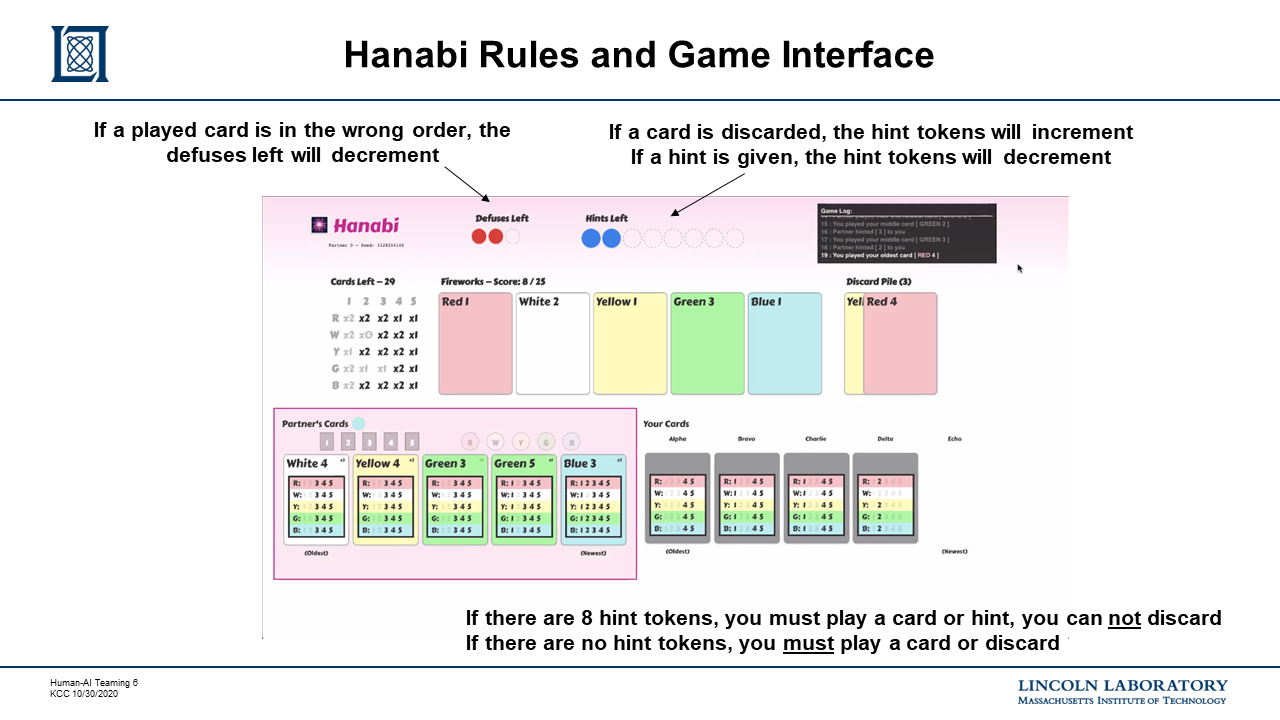}
\end{figure}
\begin{figure}[H]
    \centering
    \includegraphics[width=12cm]{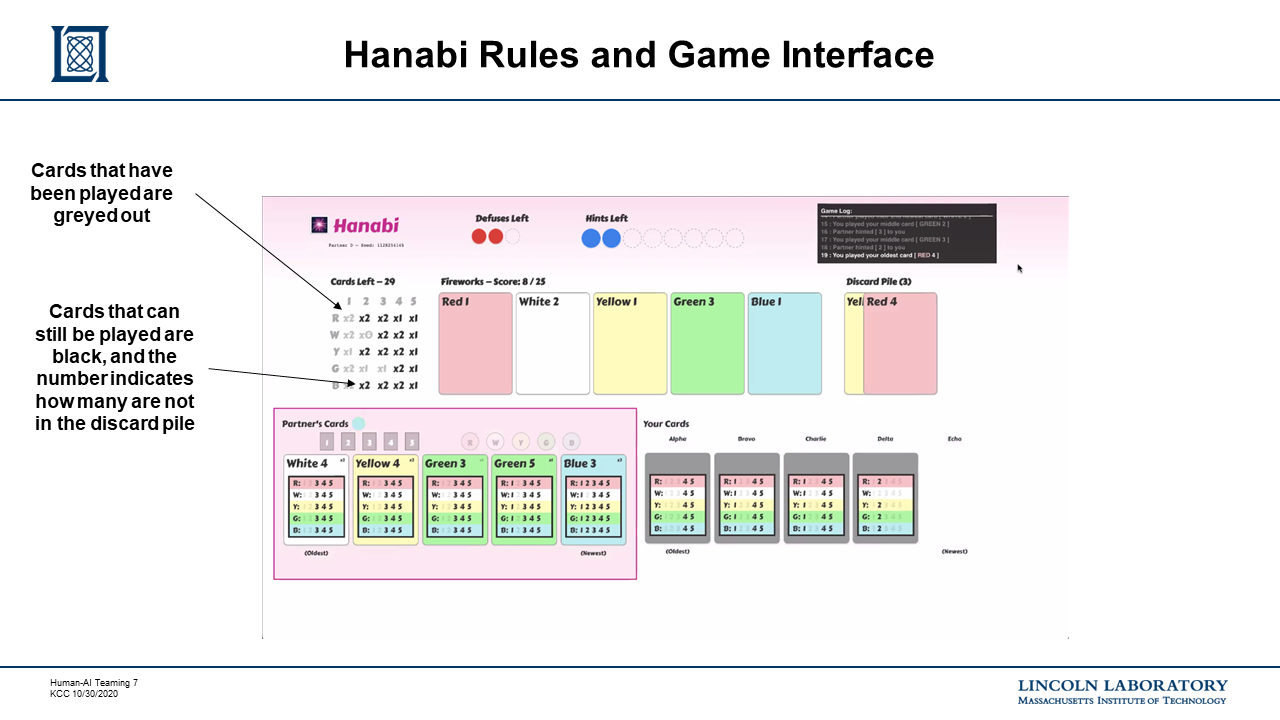}
\end{figure}
\begin{figure}[H]
    \centering
    \includegraphics[width=12cm]{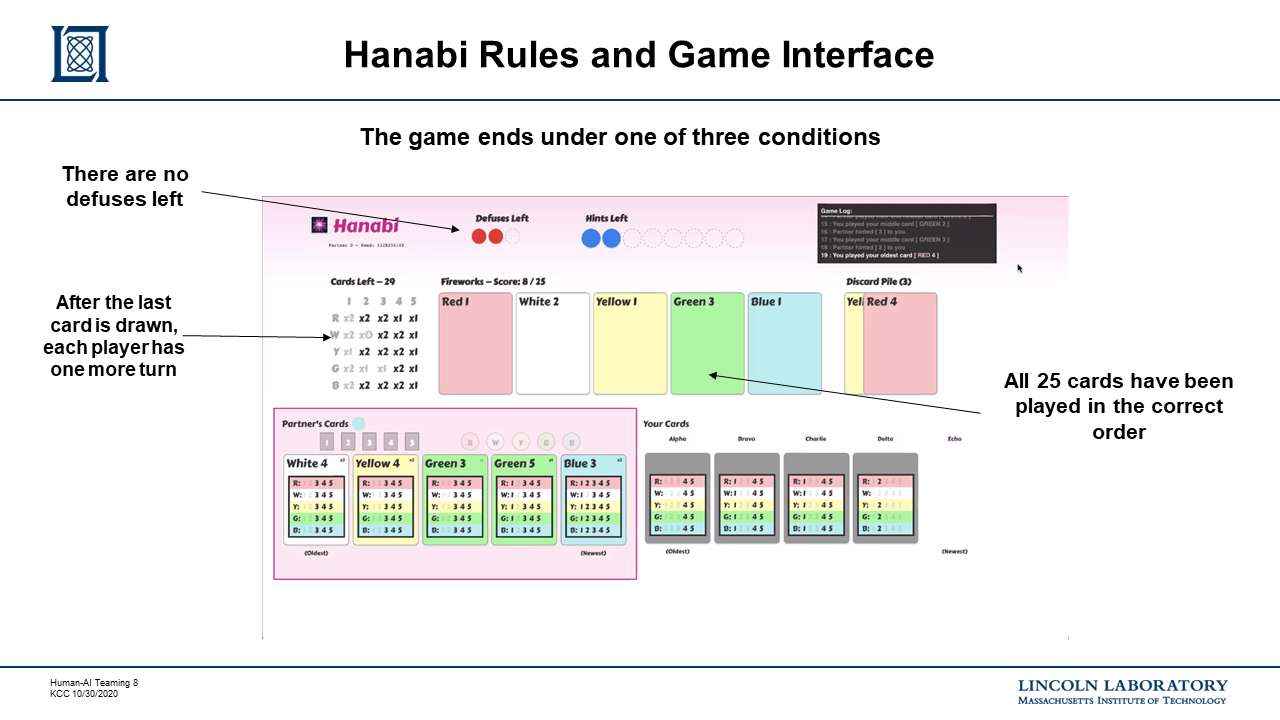}
\end{figure}

\subsection{Demographic Survey}

Questions 1-4 are two pairs of multiple choice and free response questions. Questions 5-10 are Likert scale statements with a scale from 1 (strongly disagree) to 7 (strongly agree). Questions 9 and 10 also include a free response ("Explain") if the participant indicates agreement with the statement.

% \begin{enumerate}
%     \demographicitem How often do you play card or board games? [Never, <1 hour/week, 1-3 hours/week, >3 hours/week]
%     \demographicitem Which games or types of games do you play? [free response]
%     \demographicitem How often do you play video games? [Never, <1 hour/week, 1-3 hours/week, >3 hours/week]
%     \demographicitem Which game or types of video games do you play? [free response]
%     \demographicitem I am experienced in cooperative card games.
%     \demographicitem I am experienced in cooperative board games.
%     \demographicitem I am experienced in cooperative video games.
%     \demographicitem I am experienced in Hanabi.
%     \demographicitem I am experienced in interacting with artificial intelligence agents (including voice assistants, game AIs, etc).
%     \demographicitem I am experienced in developing artificial intelligence agents.
% \end{enumerate}

\begin{table*}[ht]
\centering
\caption{Demographic Survey Prompt and Response Choices}
\label{tab:test-demo}
\small
\begin{tabular}{l|l|l}
\toprule
  & \bf{Prompt} & \bf{Response Choices}   \\ \hline
D1 & How often do you play card or board games? & [Never, <1 hour/week, 1-3 hours/week, >3 hours/week]\\
D2 & Which games or types of games do you play? & free response \\
D3 & How often do you play video games? & [Never, <1 hour/week, 1-3 hours/week, >3 hours/week] \\
D4 & Which game or types of video games do you play? & free response\\
D5 & I am experienced in cooperative card games. & Likert Scale \\
D6 & I am experienced in cooperative board games. & Likert Scale \\
D7 & I am experienced in cooperative video games. &  Likert Scale \\
D8 & I am experienced in Hanabi. &  Likert Scale\\
D9 & I am experienced in interacting with artificial & Likert Scale, free response ($optional$) \\ 
& intelligence agents (including voice \\ 
& assistants, game AIs, etc). \\
D10 & I am experienced in developing artificial & Likert Scale, free response ($optional$) \\
& intelligence agents.   \\
\end{tabular}
\end{table*}

\subsection{Demographic Survey Responses}

\begin{figure}[H]
     \centering
    \includegraphics[width=0.65\columnwidth]{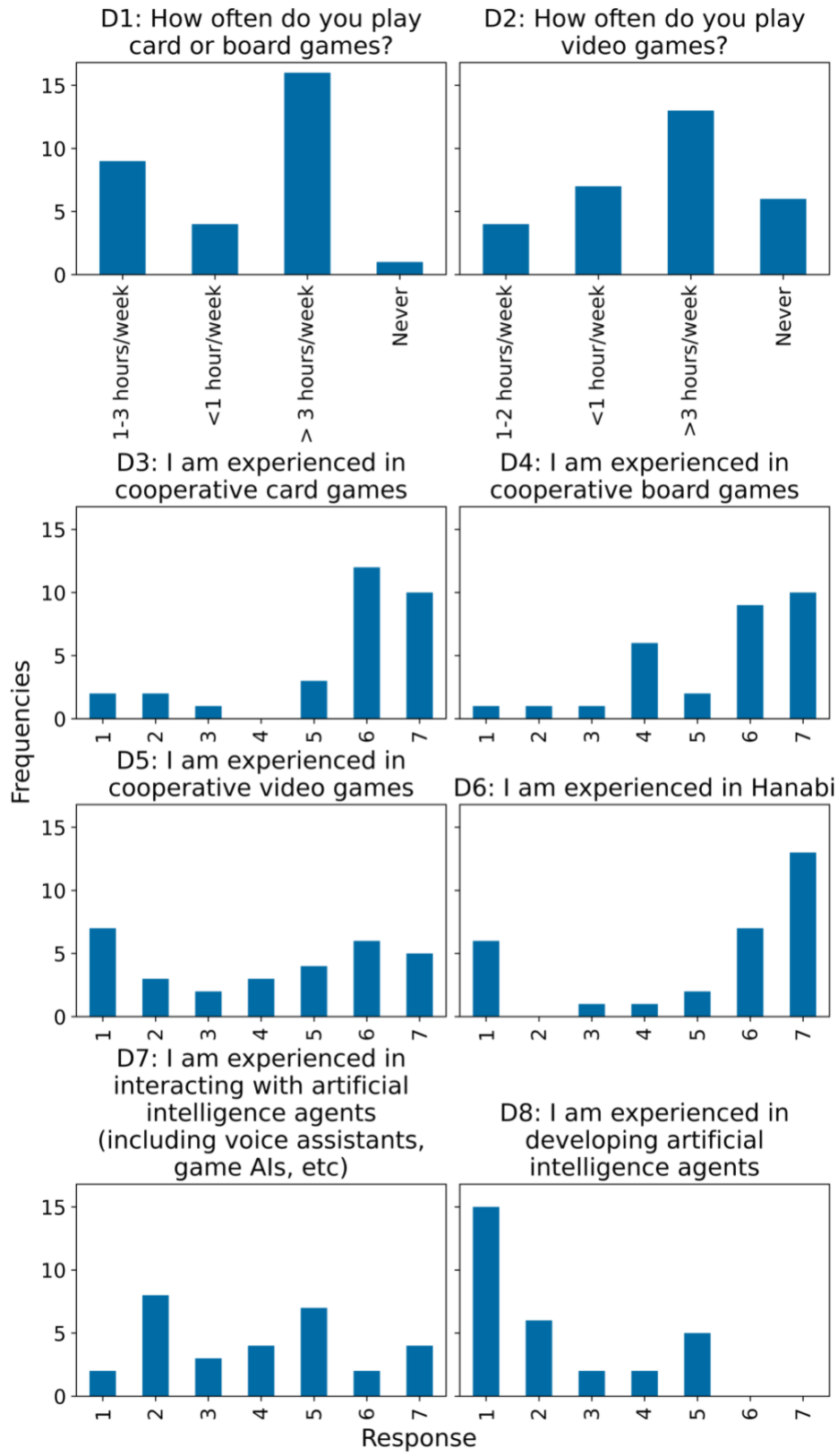}
     \caption{Histograms of all numerical and categorical demographic survey responses.}
     \label{fig:demographics_histograms}
\end{figure}

\newpage

\subsection{Post-Game Likert Scale Question Responses}

\begin{figure}[H]
     \centering
     \includegraphics[width=\columnwidth]{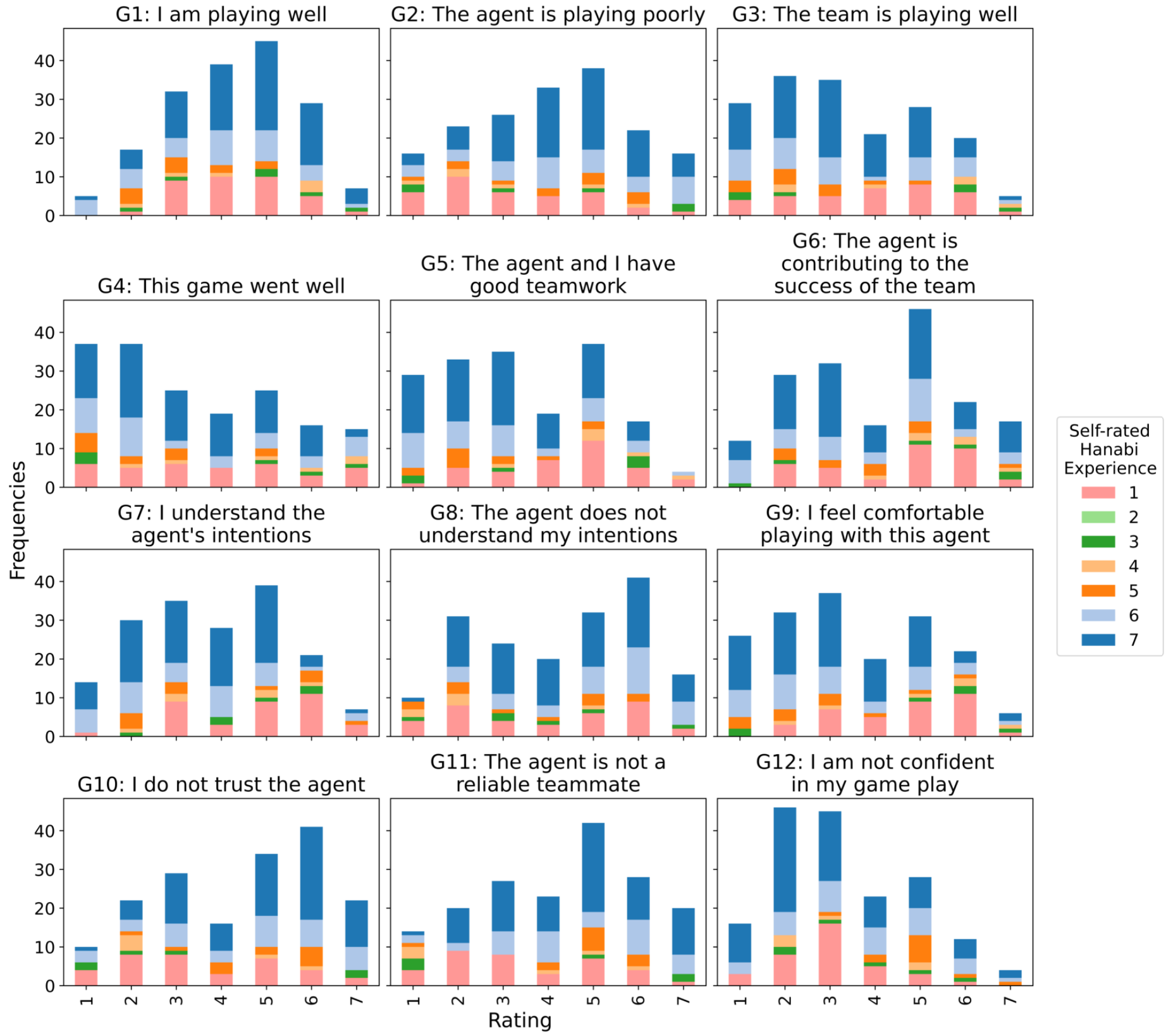}
     \caption{Participant rating for all post-game questions by self-rated Hanabi experience where statistically significant differences related to factors of agent and/or experience were presented in Section \ref{subsec:post_game_sentiments} . The scale ranges from 1-7, corresponding to "strongly disagree" to "strongly agree".}
     \label{fig:all_likert_responses}
\end{figure}

\newpage

\subsection{Participant Scores}

\begin{figure}[H]
     \centering
     \includegraphics[width=\columnwidth]{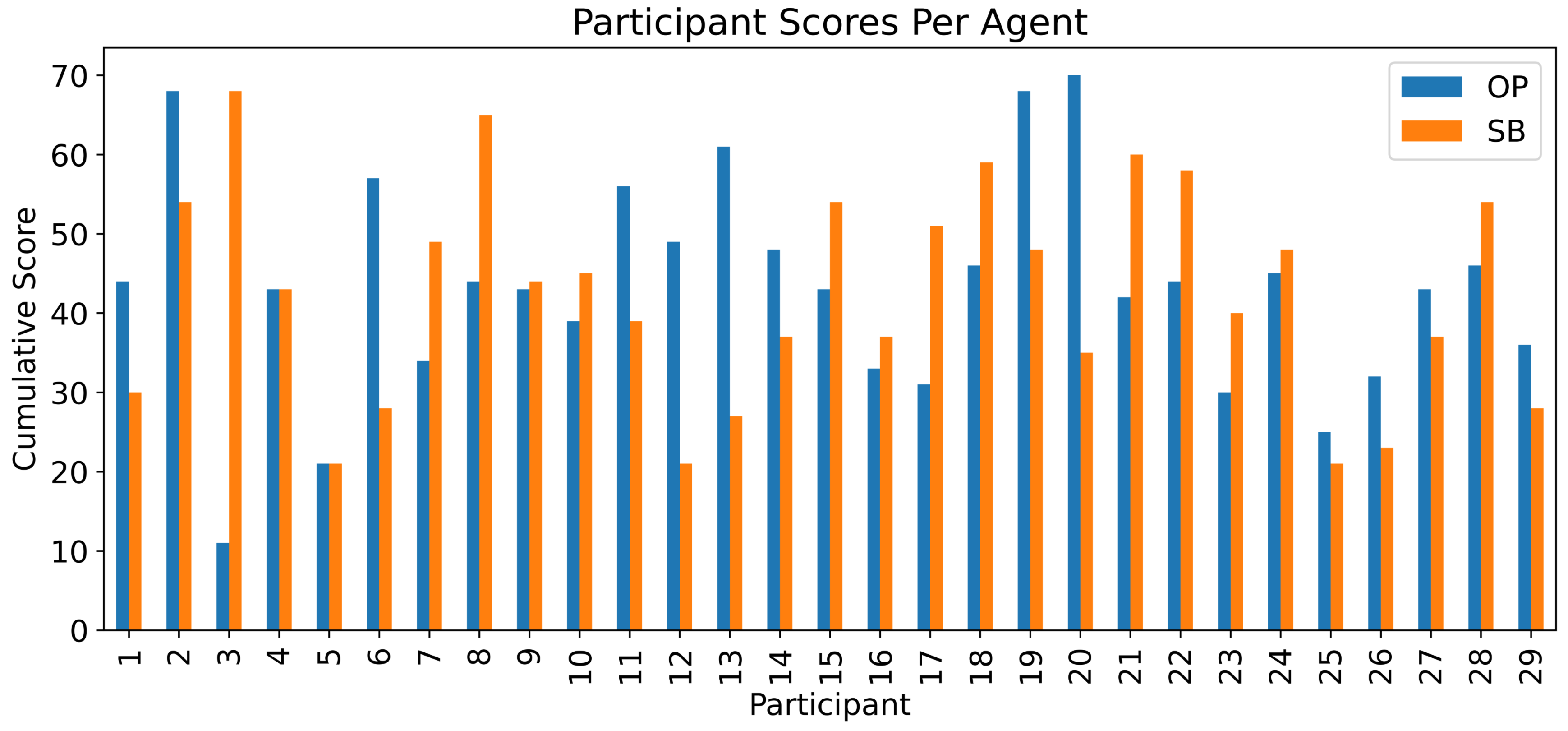}
     \caption{Cumulative game score for each participant across their six games, split by their three games with each agent type. The maximum achievable cumulative score per agent type is 75, and 150 for both. Participant 2 achieved the highest cumulative score of 122.}
     \label{fig:participant_scores}
\end{figure}

\subsection{Post-Game Survey Statistics}

Objective and subjective data were fit to second-order mixed-effects models with fixed factors of (1) AI agent, (2) self-rated Hanabi experience, (3) block (the first or second set of three games), and (4) game number (first, second, or third game within a block), and a random factor of participant number. AI agent and participant number were considered categorical variables.

\emph{Agent} refers to the AI agent type (OP or SB).

\emph{Experience} refers to the self-reported Hanabi experience level (question D8).

\emph{Lower} and \emph{Upper} are the values for the 95\% confidence intervals.

\begin{table}[H]
\caption{Model fit to response variable of game score}
\begin{tabular}{l|l|l|l|l|l|l|l}
    Name & Estimate &    SE & $t$ & df &     $p$ & Lower & Upper \\ \hline
    Agent & 1.6084 & 4.2962 & 0.37438 & 158 & 0.70863 & -6.877 & 10.094 \\
    Block & 5.282 & 2.0039 & 2.6358 & 158 & \bf{0.0092295} & 1.3241 & 9.2398 \\
    Game & 2.9572 & 1.567 & 1.8872 & 158 & 0.060961 & -0.13765 & 6.0521 \\
    Experience & 0.7736 & 0.62989 & 1.2281 & 158 & 0.22122 & -0.4705 & 2.0177\\
    Agent : Block & -3.6394 & 1.9493 & -1.867 & 158 & 0.063748 & -7.4893 & 0.21061\\
    Agent : Game & 1.1848 & 1.1962 & 0.99047 & 158 & 0.32346 & -1.1778 & 3.5473\\
    Block : Game & -0.90997 & 0.9963 & -0.91335 & 158 & 0.36245 & -2.8778 & 1.0578\\
    Agent : Experience & -0.20234 & 0.41707 & -0.48515 & 158 & 0.62824 & -1.0261 & 0.62141\\
    Block : Experience & 0.15829 & 0.36415 & 0.43468 & 158 & 0.66439 & -0.56094 & 0.87751\\
    Game : Experience & -0.13373 & 0.23366 & -0.57233 & 158 & 0.56791 & -0.59524 & 0.32777
\end{tabular}
\end{table}

\begin{table}[H]
\caption{Model fit to response variable of ``I am playing well'' (G1)}
\begin{tabular}{l|l|l|l|l|l|l|l}
    Name & Estimate &    SE & $t$ & df &     $p$ & Lower & Upper \\ \hline
    Agent &  1.9694 & 1.0546 & 1.8674 & 164 & 0.063637 & -0.11304 & 4.0518 \\
    Block &  1.9168 & 0.39977 & 4.7948 & 164 & \textbf{3.6319e-06} & 1.1274 & 2.7062 \\
    Game &  1.2204 & 0.30518 & 3.9989 & 164 & \textbf{9.6048e-05} & 0.6178 & 1.823 \\
    Experience &  0.32729 & 0.12129 & 2.6985 & 164 & \textbf{0.0076942} & 0.087808 & 0.56678 \\
    Agent : Block &  -0.80197 & 0.62076 & -1.2919 & 164 & 0.19821 & -2.0277 & 0.42375 \\
    Agent : Game &  -0.37579 & 0.20773 & -1.809 & 164 & 0.072285 & -0.78597 & 0.034391 \\
    Block : Game &  -0.39684 & 0.18093 & -2.1934 & 164 & \textbf{0.029688} & -0.75409 & -0.039593 \\
    Agent : Experience &  -0.097033 & 0.073054 & -1.3282 & 164 & 0.18595 & -0.24128 & 0.047215 \\
    Block : Experience &  -0.12775 & 0.065443 & -1.952 & 164 & 0.052636 & -0.25696 & 0.0014721 \\
    Game : Experience & -0.033548 & 0.041611 & -0.80624 & 164 & 0.42127 & -0.11571 & 0.048614 \\
\end{tabular}
\end{table}

\begin{table}[H]
\caption{Model fit to response variable of ``The agent is playing poorly'' (G2)}
\begin{tabular}{l|l|l|l|l|l|l|l}
    Name & Estimate &    SE & $t$ & df &     $p$ & Lower & Upper \\ \hline
    Agent &  -0.33608 & 1.0745 & -0.31277 & 164 & 0.75485 & -2.4577 & 1.7856 \\
    Block &  0.83106 & 0.50514 & 1.6452 & 164 & 0.10184 & -0.16635 & 1.8285 \\
    Game &  1.1238 & 0.3918 & 2.8684 & 164 & \textbf{0.0046687} & 0.35021 & 1.8975 \\
    Experience &  0.62968 & 0.15572 & 4.0436 & 164 & \textbf{8.0826e-05} & 0.3222 & 0.93717 \\
    Agent : Block &  0.45465 & 0.48308 & 0.94115 & 164 & 0.34801 & -0.49921 & 1.4085 \\
    Agent : Game &  -0.11362 & 0.29781 & -0.38151 & 164 & 0.70332 & -0.70165 & 0.47441 \\
    Block : Game &  -0.29889 & 0.24918 & -1.1995 & 164 & 0.23207 & -0.7909 & 0.19313 \\
    Agent : Experience &  0.16086 & 0.10481 & 1.5348 & 164 & 0.12677 & -0.046092 & 0.36782 \\
    Block : Experience &  -0.16477 & 0.091074 & -1.8092 & 164 & 0.072251 & -0.3446 & 0.015057 \\
    Game : Experience & -0.10097 & 0.058624 & -1.7223 & 164 & 0.086904 & -0.21672 & 0.014788 \\
\end{tabular}
\end{table}

\begin{table}[H]
\caption{Model fit to response variable of ``The team is playing well'' (G3)}
\begin{tabular}{l|l|l|l|l|l|l|l}
    Name & Estimate &    SE & $t$ & df &     $p$ & Lower & Upper \\ \hline
    Agent &  1.3927 & 0.98213 & 1.418 & 164 & 0.15807 & -0.54654 & 3.3319 \\
    Block &  2.4468 & 0.46171 & 5.2995 & 164 & \textbf{3.6985e-07} & 1.5352 & 3.3585 \\
    Game &  1.065 & 0.35812 & 2.9739 & 164 & \textbf{0.0033848} & 0.35787 & 1.7721 \\
    Experience &  -0.032564 & 0.14234 & -0.22878 & 164 & 0.81933 & -0.31361 & 0.24848 \\
    Agent : Block &  -0.58469 & 0.44155 & -1.3242 & 164 & 0.18729 & -1.4565 & 0.28716 \\
    Agent : Game &  -0.22945 & 0.2722 & -0.84293 & 164 & 0.4005 & -0.76692 & 0.30803 \\
    Block : Game &  -0.56663 & 0.22776 & -2.4879 & 164 & \textbf{0.01385} & -1.0163 & -0.11692 \\
    Agent : Experience &  -0.19952 & 0.095802 & -2.0826 & 164 & \textbf{0.03884} & -0.38868 & -0.010353 \\
    Block : Experience &  -0.076342 & 0.083244 & -0.91709 & 164 & 0.36044 & -0.24071 & 0.088027 \\
    Game : Experience & 0.051342 & 0.053584 & 0.95816 & 164 & 0.33939 & -0.054461 & 0.15715 \\
\end{tabular}
\end{table}

\begin{table}[H]
\caption{Model fit to response variable of ``The game went well'' (G4)}
\begin{tabular}{l|l|l|l|l|l|l|l}
    Name & Estimate &    SE & $t$ & df &     $p$ & Lower & Upper \\ \hline
    Agent &  0.76539 & 1.1659 & 0.65651 & 164 & 0.51242 & -1.5366 & 3.0674 \\
    Block &  2.0535 & 0.54808 & 3.7468 & 164 & \textbf{0.00024779} & 0.97132 & 3.1357 \\
    Game &  1.8013 & 0.42511 & 4.2372 & 164 & \textbf{3.7659e-05} & 0.96187 & 2.6407 \\
    Experience &  -0.16132 & 0.16896 & -0.95475 & 164 & 0.34111 & -0.49494 & 0.17231 \\
    Agent : Block &  -0.34362 & 0.52415 & -0.65558 & 164 & 0.51301 & -1.3786 & 0.69133 \\
    Agent : Game &  -0.076741 & 0.32312 & -0.2375 & 164 & 0.81257 & -0.71476 & 0.56128 \\
    Block : Game &  -0.90768 & 0.27036 & -3.3573 & 164 & \textbf{0.00097855} & -1.4415 & -0.37384 \\
    Agent : Experience &  -0.17102 & 0.11372 & -1.5038 & 164 & 0.13455 & -0.39557 & 0.053529 \\
    Block : Experience &  0.086306 & 0.098817 & 0.8734 & 164 & 0.38372 & -0.10881 & 0.28142 \\
    Game : Experience & -0.00048774 & 0.063608 & -0.0076679 & 164 & 0.99389 & -0.12608 & 0.12511 \\
\end{tabular}
\end{table}

\begin{table}[H]
\caption{Model fit to response variable of ``The agent and I have good teamwork'' (G5)}
\begin{tabular}{l|l|l|l|l|l|l|l}
    Name & Estimate &    SE & $t$ & df &     $p$ & Lower & Upper \\ \hline
    Agent &  2.3867 & 0.92567 & 2.5783 & 164 & \textbf{0.010807} & 0.55891 & 4.2145 \\
    Block &  2.5072 & 0.43517 & 5.7614 & 164 & \textbf{4.0236e-08} & 1.6479 & 3.3664 \\
    Game &  1.2232 & 0.33753 & 3.6238 & 164 & \textbf{0.00038684} & 0.55669 & 1.8896 \\
    Experience &  -0.026039 & 0.13415 & -0.1941 & 164 & 0.84634 & -0.29093 & 0.23885 \\
    Agent : Block &  -0.84574 & 0.41617 & -2.0322 & 164 & \textbf{0.043746} & -1.6675 & -0.024003 \\
    Agent : Game &  -0.36521 & 0.25656 & -1.4235 & 164 & 0.15649 & -0.87179 & 0.14137 \\
    Block : Game &  -0.59262 & 0.21467 & -2.7607 & 164 & \textbf{0.0064254} & -1.0165 & -0.16876 \\
    Agent : Experience &  -0.27275 & 0.090295 & -3.0207 & 164 & \textbf{0.0029261} & -0.45105 & -0.094464 \\
    Block : Experience &  -0.080736 & 0.078459 & -1.029 & 164 & 0.30499 & -0.23566 & 0.074185 \\
    Game : Experience & 0.024633 & 0.050504 & 0.48775 & 164 & 0.62638 & -0.075089 & 0.12435 \\
\end{tabular}
\end{table}

\begin{table}[H]
\caption{Model fit to response variable of ``The agent is contributing to the success of the team'' (G6)}
\begin{tabular}{l|l|l|l|l|l|l|l}
    Name & Estimate &    SE & $t$ & df &     $p$ & Lower & Upper \\ \hline
    Agent &  1.782 & 1.191 & 1.4963 & 164 & 0.13651 & -0.56962 & 4.1336 \\
    Block &  2.441 & 0.48735 & 5.0086 & 164 & \textbf{1.4056e-06} & 1.4787 & 3.4032 \\
    Game &  1.4781 & 0.37685 & 3.9222 & 164 & \textbf{0.00012878} & 0.73397 & 2.2222 \\
    Experience &  0.1814 & 0.14897 & 1.2177 & 164 & 0.22509 & -0.11275 & 0.47556 \\
    Agent : Block &  -0.74292 & 0.65379 & -1.1363 & 164 & 0.25747 & -2.0338 & 0.548 \\
    Agent : Game &  -0.23663 & 0.2688 & -0.88034 & 164 & 0.37997 & -0.76739 & 0.29412 \\
    Block : Game &  -0.62886 & 0.23001 & -2.7341 & 164 & \textbf{0.0069428} & -1.083 & -0.1747 \\
    Agent : Experience &  -0.17078 & 0.094562 & -1.806 & 164 & 0.072747 & -0.3575 & 0.015934 \\
    Block : Experience &  -0.070518 & 0.083572 & -0.8438 & 164 & 0.40001 & -0.23553 & 0.094498 \\
    Game : Experience & -0.065545 & 0.053426 & -1.2268 & 164 & 0.22164 & -0.17104 & 0.039946 \\
\end{tabular}
\end{table}

\begin{table}[H]
\caption{Model fit to response variable of ``I understand the agent's intentions'' (G7)}
\begin{tabular}{l|l|l|l|l|l|l|l}
    Name & Estimate &    SE & $t$ & df &     $p$ & Lower & Upper \\ \hline
    Agent &  1.4938 & 0.86912 & 1.7188 & 164 & 0.087537 & -0.22226 & 3.2099 \\
	 Block &  2.6995 & 0.40858 & 6.6069 & 164 & \textbf{5.2348e-10} & 1.8927 & 3.5062 \\
	 Game &  1.6848 & 0.31691 & 5.3164 & 164 & \textbf{3.4175e-07} & 1.0591 & 2.3106 \\
	 Experience &  -0.011884 & 0.12596 & -0.09435 & 164 & 0.92495 & -0.26059 & 0.23682 \\
	 Agent : Block &  -0.55231 & 0.39074 & -1.4135 & 164 & 0.1594 & -1.3238 & 0.21922 \\
	 Agent : Game &  -0.42332 & 0.24088 & -1.7574 & 164 & 0.08072 & -0.89895 & 0.05231 \\
	 Block : Game &  -0.78483 & 0.20155 & -3.8939 & 164 & \textbf{0.00014328} & -1.1828 & -0.38686 \\
	 Agent : Experience &  -0.19273 & 0.084778 & -2.2734 & 164 & \textbf{0.024301} & -0.36013 & -0.025335 \\
	 Block : Experience &  -0.10565 & 0.073666 & -1.4341 & 164 & 0.15343 & -0.2511 & 0.039808 \\
	 Game : Experience & 0.019661 & 0.047418 & 0.41464 & 164 & 0.67895 & -0.073968 & 0.11329 \\
\end{tabular}
\end{table}

\begin{table}[H]
\caption{Model fit to response variable of ``The agent does not understand my intentions'' (G8)}
\begin{tabular}{l|l|l|l|l|l|l|l}
    Name & Estimate &    SE & $t$ & df &     $p$ & Lower & Upper \\ \hline
    Agent &  1.1503 & 1.2751 & 0.90212 & 164 & 0.36831 & -1.3675 & 3.6682 \\
	 Block &  1.3412 & 0.51911 & 2.5838 & 164 & \textbf{0.010645} & 0.31625 & 2.3662 \\
	 Game &  1.3544 & 0.40121 & 3.3757 & 164 & \textbf{0.00091923} & 0.56216 & 2.1466 \\
	 Experience &  0.3328 & 0.15861 & 2.0982 & 164 & \textbf{0.037422} & 0.019612 & 0.64598 \\
	 Agent : Block &  -0.84719 & 0.70374 & -1.2039 & 164 & 0.23038 & -2.2367 & 0.54236 \\
	 Agent : Game &  -0.23748 & 0.28536 & -0.83219 & 164 & 0.40651 & -0.80093 & 0.32598 \\
	 Block : Game &  -0.43934 & 0.24444 & -1.7974 & 164 & 0.074115 & -0.92199 & 0.043303 \\
	 Agent : Experience &  0.3178 & 0.10039 & 3.1657 & 164 & \textbf{0.0018446} & 0.11958 & 0.51602 \\
	 Block : Experience &  -0.048055 & 0.088791 & -0.54122 & 164 & 0.58909 & -0.22338 & 0.12727 \\
	 Game : Experience & -0.10706 & 0.056744 & -1.8867 & 164 & 0.060972 & -0.2191 & 0.0049858 \\
\end{tabular}
\end{table}

\begin{table}[H]
\caption{Model fit to response variable of ``I feel comfortable playing with this agent'' (G9)}
\begin{tabular}{l|l|l|l|l|l|l|l}
    Name & Estimate &    SE & $t$ & df &     $p$ & Lower & Upper \\ \hline
    Agent &  3.143 & 0.95194 & 3.3017 & 164 & \textbf{0.0011796} & 1.2633 & 5.0226 \\
	 Block &  2.5133 & 0.44752 & 5.616 & 164 & \textbf{8.1919e-08} & 1.6296 & 3.3969 \\
	 Game &  1.0899 & 0.34711 & 3.1398 & 164 & \textbf{0.0020055} & 0.40448 & 1.7752 \\
	 Experience &  -0.023996 & 0.13796 & -0.17393 & 164 & 0.86213 & -0.29641 & 0.24841 \\
	 Agent : Block &  -1.0492 & 0.42798 & -2.4516 & 164 & \textbf{0.015273} & -1.8943 & -0.20415 \\
	 Agent : Game &  -0.34819 & 0.26384 & -1.3197 & 164 & 0.18877 & -0.86914 & 0.17277 \\
	 Block : Game &  -0.49785 & 0.22076 & -2.2552 & 164 & \textbf{0.025444} & -0.93375 & -0.061956 \\
	 Agent : Experience &  -0.33071 & 0.092857 & -3.5614 & 164 & \textbf{0.00048298} & -0.51406 & -0.14736 \\
	 Block : Experience &  -0.081598 & 0.080686 & -1.0113 & 164 & 0.31336 & -0.24092 & 0.077719 \\
	 Game : Experience & 0.022646 & 0.051937 & 0.43602 & 164 & 0.66339 & -0.079906 & 0.1252 \\
\end{tabular}
\end{table}

\begin{table}[H]
\caption{Model fit to response variable of ``I do not trust the agent'' (G10)}
\begin{tabular}{l|l|l|l|l|l|l|l}
    Name & Estimate &    SE & $t$ & df &     $p$ & Lower & Upper \\ \hline
    Agent & -0.80019 & 1.0903 & -0.73389 & 164 & 0.46406 & -2.9531 & 1.3527 \\
    Block &  1.0878 & 0.51258 & 2.1221 & 164 & \bf{0.03533} & 0.075645 & 2.0999\\
    Game & 1.1899 & 0.39758 & 2.9929 & 164 & \bf{0.0031911} & 0.40487 & 1.9749 \\
    Experience & 0.6055 & 0.15802 & 3.8318 & 164 & \bf{0.00018089} & 0.29349 & 0.91752 \\
    Agent : Block & 0.83804 & 0.4902 & 1.7096 & 164 & 0.089233 & -0.12987 & 1.806\\
    Agent : Game & -0.20681 & 0.3022 & -0.68436 & 164 & 0.49472 & -0.80351 & 0.38989\\
    Block : Game & -0.37637 & 0.25285 & -1.4885 & 164 & 0.13854 & -0.87564 & 0.1229\\
    Agent : Experience & 0.203 & 0.10636 & 1.9086 & 164 & 0.058055 & -0.0070082 & 0.413\\
    Block : Experience & -0.17691 & 0.092417 & -1.9143 & 164 & 0.057328 & -0.35939 & 0.0055706\\
    Game : Experience & -0.076328 & 0.059488 & -1.2831 & 164 & 0.20127 & -0.19379 & 0.041133
\end{tabular}
\end{table}

\begin{table}[H]
\caption{Model fit to response variable of ``The agent is not a reliable teammate'' (G11)}
\begin{tabular}{l|l|l|l|l|l|l|l}
    Name & Estimate &    SE & $t$ & df &     $p$ & Lower & Upper \\ \hline
    Agent & 0.34397 & 1.1345 & 0.3032 & 164 & 0.76212 & -1.8961 & 2.584\\
    Block & 1.0037 & 0.53332 & 1.8821 & 164 & 0.061598 & -0.049315 & 2.0568\\
    Game & 1.0982 & 0.41366 & 2.6549 & 164 & \bf{0.0087148} & 0.28145 & 1.915\\
    Experience & 0.51945 & 0.16441 & 3.1594 & 164 & \bf{0.0018826} & 0.19481 & 0.84409\\
    Agent : Block & 0.12847 & 0.51003 & 0.25188 & 164 & 0.80145 & -0.87861 & 1.1355\\
    Agent : Game & -0.1499 & 0.31442 & -0.47675 & 164 & 0.63417 & -0.77074 & 0.47094\\
    Block : Game & -0.30352 & 0.26308 & -1.1537 & 164 & 0.2503 & -0.82299 & 0.21595\\
    Agent : Experience & 0.12813 & 0.11066 & 1.1579 & 164 & 0.24859 & -0.090369 & 0.34664\\
    Block : Experience & -0.088578 & 0.096156 & -0.9212 & 164 & 0.3583 & -0.27844 & 0.10128\\
    Game : Experience & -0.097645 & 0.061895 & -1.5776 & 164 & 0.11659 & -0.21986 & 0.024569
\end{tabular}
\end{table}

\begin{table}[H]
\caption{Model fit to response variable of ``I am not confident in my gameplay'' (G12)}
\begin{tabular}{l|l|l|l|l|l|l|l}
    Name & Estimate &    SE & $t$ & df &     $p$ & Lower & Upper \\ \hline
    Agent &  1.4796 & 1.2583 & 1.1758 & 164 & 0.24137 & -1.005 & 3.9641 \\
	 Block &  1.5481 & 0.4575 & 3.3839 & 164 & \textbf{0.00089398} & 0.64478 & 2.4515 \\
	 Game &  0.61849 & 0.34399 & 1.798 & 164 & 0.074017 & -0.060728 & 1.2977 \\
	 Experience &  0.26304 & 0.13801 & 1.906 & 164 & 0.058405 & -0.0094638 & 0.53554 \\
	 Agent : Block &  -1.1132 & 0.76425 & -1.4565 & 164 & 0.14716 & -2.6222 & 0.39588 \\
	 Agent : Game &  0.037263 & 0.22698 & 0.16417 & 164 & 0.8698 & -0.41092 & 0.48544 \\
	 Block : Game &  -0.29982 & 0.20014 & -1.498 & 164 & 0.13605 & -0.69501 & 0.095368 \\
	 Agent : Experience &  0.031346 & 0.079802 & 0.39279 & 164 & 0.69498 & -0.12623 & 0.18892 \\
	 Block : Experience &  -0.044646 & 0.07217 & -0.61861 & 164 & 0.53703 & -0.18715 & 0.097857 \\
	 Game : Experience & -0.048831 & 0.045717 & -1.0681 & 164 & 0.28704 & -0.1391 & 0.041439 \\
\end{tabular}
\end{table}

\subsection{Novice vs Expert Post-Game $t$-Tests}

Post-hoc pairwise comparisons of novice vs expert in cases where agent and self-reported Hanabi experience have significant interaction effects, as described in Section \ref{subsec:post_game_sentiments}.

\begin{table}[H]
\centering
\caption{Two-sample $t$-tests of post-game sentiment, comparing novice and expert reactions.}
\begin{tabular}{l|l|l|l|l|l}
    Question & Agent & $t$ & $p$ & corrected $p$ & $d$\\ \hline
    G5 & SB & 0.35599 & 0.72273 & 1.00000 & 0.080708 \\
    G5 & OP & 5.1395 & 1.7334e-06 & \textbf{1.38672e-05} & 1.0185 \\
    G9 & SB & 0.25536 & 0.79906 & 1.00000 & 0.057915 \\
    G9 & OP & 5.8552 & 8.7246e-08 & \textbf{7.85214e-07} & 1.1214 \\
    G8 & SB & 0.61126 & 0.54266 & 1.00000 & 0.13838 \\
    G8 & OP & -5.9229 & 6.5231e-08 & \textbf{6.52310e-07} & -1.1306 \\
    G3 & SB & -1.1856 & 0.2391 & 0.956400 & -0.26679 \\
    G3 & OP & 3.5514 & 0.00062779 & \textbf{3.76674e-03} & 0.75189 \\
    G7 & SB & 1.652 & 0.10223 & 0.511150 & 0.36893 \\
    G7 & OP & 5.0678 & 2.3171e-06 & \textbf{1.62197e-05} & 1.0076
\end{tabular}
\end{table}

\subsection{Post-Experiment $t$-Tests}

One-sample $t$-tests of post-experiment sentiment. Some responses were flipped on the Likert scale for directional consistency, based on which agent was seen first, since the ends of the scale were labeled as the ``first'' and ``second'' agent for the participants. Preference directionality is such that 1 is towards OP and 7 is towards SB. $t$ statistics greater than 0 indicate answering towards SB. The Holm–Bonferroni step-down method was used for multiple comparisons correction. $d$ is the Cohen's effect size. In general, thresholds for ``small,'' ``medium,'' and ``large'' effect sizes are considered to be $|d| = 0.2, 0.5,$ and $0.8$ respectively \cite{cohen1977statistical}.

\begin{table}[H]
\centering
\caption{One-sample $t$-tests of post-experiment sentiment.}
\begin{tabular}{l|l|l|l|l}
    Question & $t$ & $p$ & corrected $p$ & $d$\\ \hline
    Which agent did you prefer playing with?  &  2.90633  &  0.00707 & \textbf{0.03969} & 0.549\\
    Which agent did you trust more?  &  3.40564  &  0.00201 & \textbf{0.01610} & 0.644\\
    Which agent did you understand more?  &  2.88618  &  0.00743 & \textbf{0.03969} & 0.545\\
    Which agent understood you better?  &  2.93369  &  0.00661 & \textbf{0.03969} & 0.554\\
    Which agent was the better Hanabi player?  &  3.36011  &  0.00226 & \textbf{0.01610} & 0.635\\
    Which agent was more reliable?  &  2.86217  &  0.00788 & \textbf{0.03969} & 0.541\\
    Which agent had a better understanding of &&&\\
    the game on average?  &  2.68186  &  0.01214 & \textbf{0.03969} & 0.507\\
    Which agent caused you to have a greater &&&\\
    mental workload?  &  -0.16385  &  0.87103 & 0.87103 & -0.031
\end{tabular}
\end{table}

\subsection{Post-Experiment Participant Preference and Free Response}

Post-experiment ratings of agent preference and explanation of the preference. The ``Preference'' heading corresponds to a Likert-scale response to the question ``Which agent did you prefer playing with?'' where 1 was ``the first agent'' and 7 was ``the second agent.'' ``Explanation'' was a free-response field with the question ``Why did you prefer the agent that you did?''

\begin{longtable}{l|l|l|p{0.6\linewidth}}
\caption{Post-experiment ratings of agent preference and explanation}\\

Participant & Order & Preference & Explanation \\ \hline
1 &	SB, OP &	5 &	first agent felt like it was learning; really bad to begin with; had to "teach" them how to play hanabi; second agent felt like someone who knew how to play hanabi and wanted to trick you; broke my trust in 2nd game; in 3rd game was "trust me"; don't like playing with 2nd agent.\\ \hline
2 &	OP, SB&	6&	It seemed to have a better understanding of not just what hints to give, but when to give them. I think a lot of the strategies cascaded from that - both my strategy and its. It just had a better understanding of the tempo of the game. If you think of it as - every time I get a hint, I have to perform an MLE, and when I give a hint, that's what they have to perform - if you just take the clue at face value; you can think about "why am I giving this hint" the second agent thought about "why am I getting/giving this hint NOW" while the first agent didn't.\\\hline
3&	SB, OP&	1&	It knew the rules of the game. It knew how to play.\\ \hline
4&	OP, SB&	4&	I thought the first one was dumber but more consistent. The second one - I thought I was starting to understand it in the second game, but then in the third game, I completely didn't understand what it was doing at all.\\ \hline
5&	SB, OP&	1&	Second agent made an obvious mistake quite frequently. There are some cases where it was clear that if the agent played a card, we would lose the game, but it played it anyway. Sometimes it would also give me hints that I already know. \\ \hline
6&	OP, SB&	7&	better able to understand what it's clues meant and how to give it clues that would result in the correct actions\\ \hline
7&	SB, OP&	3&	The first agent was more predictable, even if I didn't necessarily agree with their strategy. Both of them made dumb choices, like playing cards that were clearly not playable when they had full information on them (or at least enough information), or they discarded cards with full information and were playable.\\ \hline
8&	OP, SB&	7&	The second agent seemed to be more capable of inductive reasoning than the first. Both has similar styles of inductive clues, but it seemed like the second took inductive clues better. The discard strategy of the first agent felt worse than the discard strategy of the second.\\ \hline
9&	SB, OP&	1&	agent 1 was more consistent; even if i didn't understand what they were doing, i could more reliably assume they would play or discard cards if they knew they were playable; I feel I bombed the second one whenever i clued it; did not know how it would react\\ \hline
10&	OP, SB&	6&	Maybe it's because the first one was so terrible that it made me have zero expectation of the second one. So even though the second agent wasn't that much better, and I was confused by its strategy, I was used to being confused and wasn't surprised anymore. It took less willpower to go through the games [with the second agent].\\ \hline
11&	SB, OP&	7&	Gave me info; seemed to act on cues better; it felt like there was 2-way comms as opposed to 1-way; also it didn't throw away cards (e.g., knew perfect info on)\\ \hline
12&	OP, SB&	7&	It provided more challenge and interest. Because I could reasonably play with it. It let me play at a more satisfying level.\\ \hline
13&	SB, OP&	1&	Agent 1 seemed to have a better model of the game in the sense that it deliberately played playable cards and discarded unplayable cards more frequently; as opposed to the 2nd agent that played known unplayable cards and did not play cards when it had the chance to; first agent was more inline with game of Hanabi rules of playing all cards when possible; first agent played a way that was more familiar; First agent still used strategies that were more human friendly; i understood it better and it understood me better\\ \hline
14&	OP, SB&	3&	I felt like the first agent was improving and started understanding my strategy more, whereas the second one wasn't learning from the errors or mistakes that both of us made.\\ \hline
15&	SB, OP&	6&	The rules that the second (2nd) agent was following was easier to understand; specifically the discarding strategy was much more predictable; first agent may have predictable discarding strategy, but the 2nd agent is much easier to play with\\ \hline
16&	OP, SB&	7&	I probably had some learning effects for the game so I understood things better, however, I also found that it was easier to get into a cadence of play with the second agent. I think I understood the intention of the second agent and it understood me.\\ \hline
17&	SB, OP&	1&	The first agent played cards and hinted cards consistently. The second agent by contrast did not play multiply-hinted cards and gave hints that were not necessarily playable. With the first agent, I could reasonably expect to perform well and to trust his decision whereas with the second agent, I found myself trying to compensate for his lack of reliability.\\ \hline
18&	OP, SB&	2&	i was better at predicting what the first agent would do; after first 2 games i understood the agent's strategy though i didn't agree with it; with 2nd agent i couldn't figure out how it's saves and discards worked and that made it impossible for me to tell it to save cards i wanted to protect\\ \hline
19&	SB, OP&	1&	because i could understand what it would do and i can predict what they would do better; and i have opinion that i can understand the clue of the first agent and what the agent tries to force me to do; the first agent preferred to play instead of discarding; second agent prefers to discard instead of play which is sub-optimal in hanabi game (i.e., it had full info about a card and still chose to discard)\\ \hline
20&	OP, SB&	7&	It does understand rules of Hanabi among humans.\\ \hline
21&	SB, OP&	7&	It gave me more hints and it didn't make inexplicable discard decisions that were clearly suboptimal based on information that it had at the time. It was also the only one of the six games that we completed (25 pts).\\ \hline
22&	OP, SB&	6&	second agent played color clues that i gave\\ \hline 
23&	SB, OP&	5&	I felt that the second agent understood some clues better than the first even though i think they are very very similar; similar strategy on saving discarding cluing; main difference for the second one was that it would clue sooner than the first one; it wouldn't delay cluing even though it had cards; 3 games is a bit short to determine/assess strategy; \\ \hline
24&	OP, SB&	6&	The main reason was that Agent 2 was willing to change its discard behavior to match mine, as I strongly prefer discarding the oldest card instead of the newest. The other reason is the second agent was a little better at giving clues to me that I understood the meaning of.\\ \hline
25&	SB, OP&	7&	It seemed more cooperative in that it was giving a lot of hints and it seemed like we had a similar strategy. Early on, we'd tell each other when we had ones, and then giving full information, giving the appropriate hints for the state of the game. Seemed like we had good teamwork. They were giving hints, and also taking hints.\\ \hline
26&	OP, SB&	7&	The second agent understood my strategy better; it was easier for me to follow it's pattern/strategy; and because we got closer to winning, ergo, it was doing something right\\ \hline
27&	SB, OP&	2&	the first agent provides more certainty even though the game progresses slower, it acts upon certainty and minimizes guessing;\\ \hline
28&	OP, SB&	6&	i felt like the second agent was playing with easier to understand set of rules; they appeared to be more mindful of hints or number of hints remaining, so there is a better back-and-forth depending on who what playable cards or not; \\ \hline
29&	SB, OP&	3&	To my understanding, the strategy seemed very consistent and simple. Agent 2's strategy seemed more complex and less predictable. It seemed more random which is less preferable.
\end{longtable}

\end{document}